\documentclass{article}

\usepackage[
  backend   = biber,
  style     = numeric,   
  sorting   = none,      
  maxcitenames = 99,     
  maxbibnames = 99       
]{biblatex}
\usepackage[final,nonatbib]{neurips_2025}

\usepackage[utf8]{inputenc} 
\usepackage[T1]{fontenc}    
\usepackage{hyperref}       
\usepackage{url}            
\usepackage{booktabs}       
\usepackage{amsmath}       
\usepackage{amsfonts}       
\usepackage{nicefrac}       
\usepackage{microtype}      
\usepackage{xcolor}         
\usepackage{xspace}
\usepackage{times}
\usepackage{latexsym}
\usepackage{soul}
\usepackage{tabularx,colortbl}
\usepackage{tabularx}
\usepackage{makecell}
\usepackage{multirow}
\usepackage{float}
\usepackage{arydshln} 
\usepackage{wrapfig}
\usepackage{graphics}
\usepackage{graphicx}
\usepackage{algpseudocode}
\usepackage{listings}
\usepackage{amssymb}
\usepackage{algorithm}
\usepackage{enumitem}
\usepackage{tcolorbox}
\usepackage{subcaption}

\newcommand{\ours}{{Zebra-Llama}\xspace}
\newcommand{\smart}{{SMART}\xspace}

\title{\begin{wrapfigure}{L}{0.12\textwidth}
    \vspace{-0.4in}
    \hspace{-110pt}
    \includegraphics[width=0.09\textwidth]{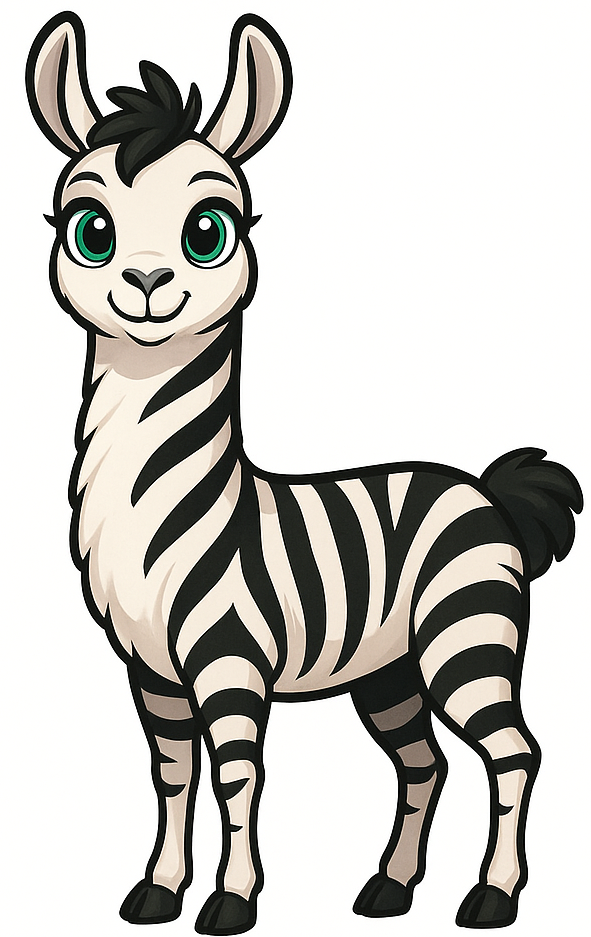}
    \vspace{-0.08in}  
    \vspace{-1.5mm}
\end{wrapfigure} 
\hspace{-0.5em} 
\ours: Towards Extremely Efficient Hybrid Models
\vspace{-10pt}}

\author{
Mingyu Yang$^*$, Mehdi Rezagholizadeh$^*$, Guihong Li\thanks{ Equal Contribution First Authors, with order determined randomly. } , Vikram Appia, Emad Barsoum \\
Advanced Micro Devices, Inc. (AMD)\\
\texttt{\{mingyu.yang,mehdi.rezagholizadeh,guihong.li\}@amd.com} \\
}

\begin{document}

\maketitle
\vspace{-15pt}
\begin{abstract}
\vspace{-8pt}
With the growing demand for deploying large language models (LLMs) across diverse applications, improving their inference efficiency is crucial for sustainable and democratized access. However, retraining LLMs to meet new user-specific requirements is prohibitively expensive and environmentally unsustainable. In this work, we propose a practical and scalable alternative: composing efficient hybrid language models from existing pre-trained models. 
Our approach, \textit{\ours}, introduces a family of 1B, 3B, and 8B hybrid models by combining State Space Models (SSMs) and Multi-head Latent Attention (MLA) layers, using a refined initialization and post-training pipeline to efficiently transfer knowledge from pre-trained Transformers.
\ours achieves Transformer-level accuracy with near-SSM efficiency using only 7--11B training tokens (compared to trillions of tokens required for pre-training) and an 8B teacher. Moreover, \ours dramatically reduces KV cache size---down to 3.9\%, 2\%, and 2.73\% of the original for the 1B, 3B, and 8B variants, respectively---while preserving 100\%, 100\%, and $>$97\% of average zero-shot performance on LM Harness tasks. Compared to models like MambaInLLaMA, X-EcoMLA, Minitron, and Llamba, \ours consistently delivers competitive or superior accuracy while using significantly fewer tokens, smaller teachers, and vastly reduced KV cache memory. Notably, \ours-8B surpasses Minitron-8B in few-shot accuracy by 7\% while using $8\times$ fewer training tokens, over $12\times$ smaller KV cache, and a smaller teacher (8B vs. 15B). It also achieves {$1.4\times$--$3.3\times$} higher throughput (tokens/s) than MambaInLlama. {The source code is released at \url{https://github.com/AMD-AGI/AMD-Hybrid-Models}.} 
\end{abstract}

\vspace{-4.5mm}
\section{Introduction}
\vspace{-3mm}
\begin{wrapfigure}{R}{0.45\textwidth}
    \centering
    \vspace{-0.3in}
    \includegraphics[width=0.43\textwidth]{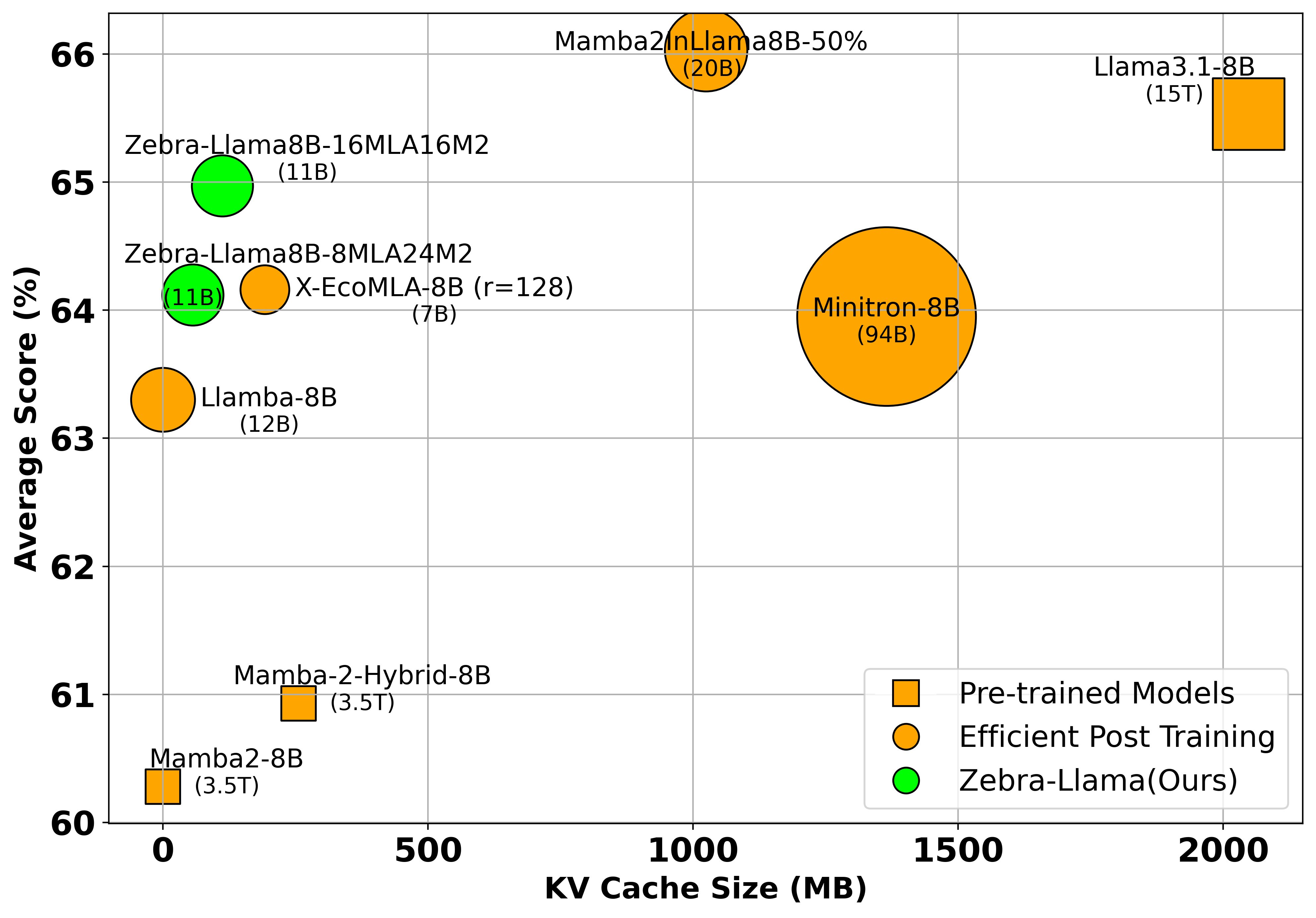}
    \vspace{-0.08in}
    \caption{\small{Comparing 8B-scale models on average LM Harness score vs. KV cache size. \ours (green) matches or exceeds baselines with smaller KV cache and fewer training tokens. Circle and square sizes indicate training tokens (billions for post-training, trillions for pre-training).}}  
    \label{fig:sensitivity}
    \vspace{-1.5mm}
\end{wrapfigure}
The exponential growth of deep learning applications has created an urgent demand for models that strike a balance between accuracy and computational efficiency—particularly in scenarios constrained by memory or limited hardware capabilities. Transformer-based models, despite their impressive performance across a range of tasks, are fundamentally limited by the quadratic complexity of their self-attention mechanisms and the substantial memory required to store key–value (KV) caches. These bottlenecks hinder their deployment in real-world applications, especially on edge devices or in latency-sensitive settings.
At the same time, the rise of large language models (LLMs) has amplified the need for \emph{customization}—that is, the ability to adapt pre-trained models to meet diverse user needs, hardware configurations, and application requirements. However, developing new LLMs from scratch for each target environment is prohibitively expensive and environmentally unsustainable. Traditional solutions such as model compression, neural architecture search (NAS), and pre-training new architectures offer potential pathways but suffer from significant limitations. Model compression often degrades quality while NAS or pre-training new models have substantial computational costs.

To overcome these challenges, a promising new paradigm has emerged: {hybrid models} that aim to reduce the computational cost of self-attention while maintaining generation quality. These architectures typically integrate efficient state-space models~\cite{gu_mamba_nodate} or linear attention~\cite{katharopoulos2020transformers,} with full attention mechanisms, leveraging the strengths of both. However, many recent hybrid approaches, including Samba~\cite{ren2024samba} and Hymba~\cite{dong2024hymba}, require extensive pre-training from scratch, which is computationally expensive. Others, such as MambaInLLaMA~\cite{wang2025mamballamadistillingaccelerating}, experience notable performance degradation when attention is replaced too aggressively, often due to insufficient key–value cache compression or ineffective knowledge transfer from the base model.

{Our goal in this work is to develop a more efficient and sustainable alternative: to \emph{compose} highly efficient language models directly from existing pre-trained Transformers}, avoiding the cost of full pre-training while retaining performance. 
Our approach, called \textit{\ours}, introduces a family of hybrid models (1B, 3B, and 8B) built on two complementary components: Multi-Latent Attention (MLA)~\cite{liu2024deepseek}, a low-rank attention mechanism that compresses memory usage without sacrificing quality under moderate compression; and Mamba2, a state-space model that eliminates KV caches entirely but performs poorly when used alone~\cite{wang2025mamballamadistillingaccelerating}. 
Specifically, we first initialize pure MLA and pure Mamba2 models from a pre-trained Transformer via a refined weight-mapping procedure. We then use Intermediate Layer Distillation (ILD) to align their internal representations with those of the original Transformer model, ensuring strong initialization. Finally, we strategically compose hybrid architectures from the refined MLA and Mamba2 variants using a sensitivity-aware strategy called \textbf{SMART} (Sensitivity Measure-Aware Replacement of Transformer layers) to select where each component is most effective. This process results in highly efficient models that retain Transformer-level quality with drastically reduced memory and compute requirements.
Our \ours family of 1B, 3B, and 8B hybrid models, achieving {$25\times$, $50\times$, and $36\times$} KV cache compression relative to their respective base Transformer models, while maintaining \textbf{100\%}, \textbf{100\%}, and \textbf{>97\%} of the base model’s average zero-shot performance on the LM Harness evaluation benchmark. 
Our models also perform competitively on few-shot tasks. Notably, \ours-8B improves the average few-shot accuracy over Minitron-8B by \textbf{7\%}, despite using \textbf{$8\times$} fewer training tokens, \textbf{>12$\times$} smaller KV cache, and a smaller teacher model (8B vs. 15B). Additionally, our \ours exhibits high inference speed. Compared to the existing work MambaInLlama, it achieves {\textbf{1.4--3.3$\times$}} higher throughput {with 1.8-5.5$\times$ less memory consumption}. 
This work introduces a practical route for building efficient, customizable LLMs from existing models.
 Our key contributions are:
\vspace{-2mm}
\begin{itemize}[leftmargin=10pt, itemsep=0pt]
\item \textbf{Architecture:} We propose a hybrid model combining MLA and Mamba2 layers, replacing classical Transformer blocks to reduce memory usage and address the quadratic bottleneck of attention.
\item \textbf{Training:} We develop an efficient post-training pipeline including refined initialization, Intermediate Layer Distillation, and SMART layer selection for hybrid model composition.
\item \textbf{Empirical Results:} Our models match or exceed Transformer-level performance with drastically reduced KV cache and significantly improved inference throughput.
\end{itemize}

\vspace{-4mm}
\section{Related Work}
\vspace{-3mm}
\paragraph{Hybrid Models}
Prior work on hybrid models can broadly be categorized into two groups based on their training strategy: \emph{pre-training-based} and \emph{post-training-based} approaches. Pre-training-based methods, such as Jamba~\cite{lieber2024jamba}, Hymba~\cite{dong2024hymbahybridheadarchitecturesmall}, Samba~\cite{ren2024samba} and Mamba-2-Hybrid~\cite{waleffe2024empirical}, interleave heterogeneous hybrid layers during full-scale model training, allowing the development of hybrid models from scratch. While these models achieve strong performance, their training cost remains high, limiting accessibility and sustainability. In contrast, post-training approaches insert efficient modules into pre-trained Transformers, often leveraging knowledge distillation to transfer capabilities without full re-training, including MambaInLLaMA~\cite{wang2025mamballamadistillingaccelerating}, MOHAWK~\cite{bick2024transformers}, and Llamba~\cite{bick2025llamba}.
Our work focuses on the post-training setting and pushes it further by targeting extremely efficiency—both in terms of training tokens and runtime memory. We introduce a systematic method to build hybrid models with minimal training costs, making them practical for broad deployment, especially where hardware resources are critically constrained.

\vspace{-3mm}
\paragraph{Efficient Post-training}
Several recent approaches specifically target post-training efficiency through careful initialization, distillation, and model compression. For example, MambaInLLaMA~\cite{wang2025mamballamadistillingaccelerating} replaces most self-attention blocks of a pre-trained Transformer with linear RNN layers. Through the initialization of RNN layer parameters from the pre-trained Transformer's weights, followed by distillation-based fine-tuning, the hybrid model achieves comparable performance with the teacher model. MOHAWK~\cite{bick2024transformers} introduces a multi-stage cross-architecture distillation strategy (aligning attention “mixing” matrices, hidden states, and outputs) to transfer a Transformer’s knowledge into a Mamba-based model using a fraction of the original training data. Building on this, Llamba~\cite{bick2025llamba} scales the distilled recurrent architecture up to 8B parameters, attaining improved accuracy, while markedly improving inference throughput and memory usage for deployment on edge devices. Orthogonally, X-EcoMLA~\cite{li2025x} “upcycles” a pre-trained Transformer’s attention into multi-head latent attention modules, jointly compressing key–value caches significantly with teacher-guided fine-tuning to preserve accuracy. Meanwhile, Minitron~\cite{muralidharan2024compact} compresses a 15B Nemotron Transformer by pruning its depth, width, and attention heads, then retrains on only about 2–3\% of the original data via distillation; this yields 8B and 4B models that rival larger models’ performance without full re-training. Our method offers significant advantages over previous distillation techniques like {Minitron}, MambaInLlama, X-EcoMLA, and Llamba, which have each faced limitations in KV cache compression, training efficiency or maintaining the base models' performance.


\begin{figure}
    \centering
    \includegraphics[width=0.99\linewidth]{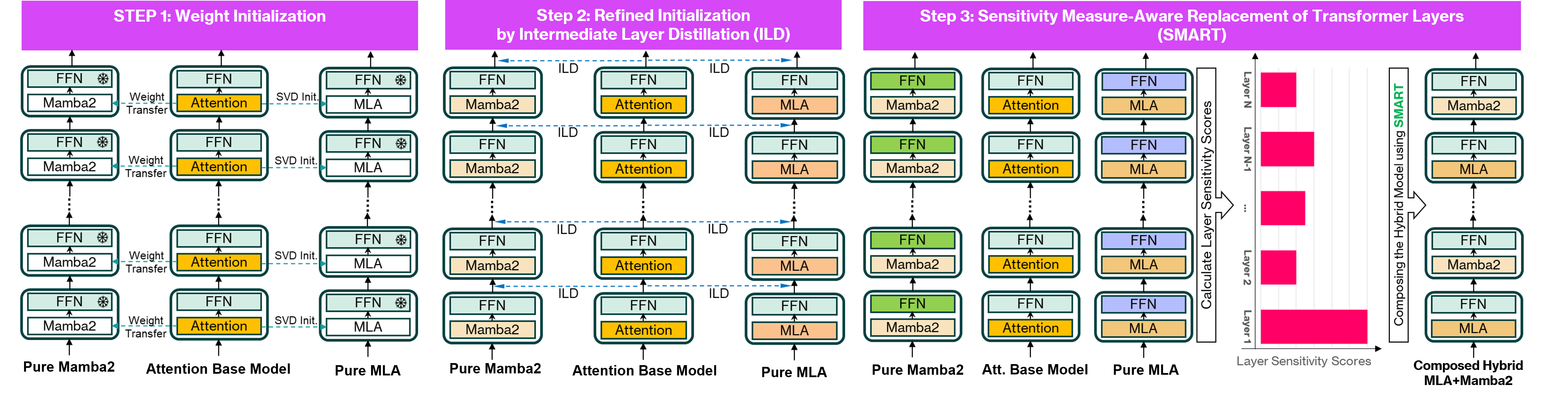}
    \vspace{-2mm}
    \caption{\small{Overview of our hybrid model composition pipeline. The process consists of three stages: (1) \textbf{Weight Initialization} -- we initialize pure Mamba2 and MLA models from a pre-trained Transformer via structured mapping; (2) \textbf{Refined Initialization through Intermediate Layer Distillation (ILD)} -- we refine both models by aligning their internal representations with the base model on a small dataset; and (3) \textbf{SMART Layer Selection} -- we compose the final hybrid model by selecting MLA and Mamba2 layers based on sensitivity analysis. }}\label{fig:overview}
    \vspace{-5mm}
\end{figure}

\vspace{-8pt}
\section{Methodology}
\vspace{-8pt}
Our methodology focuses on designing and training a hybrid model architecture that achieves strong performance with significantly enhanced efficiency. The overall approach consists of two key stages: (1) constructing an extremely efficient hybrid model, and (2) applying a lightweight yet effective training pipeline. An overview of the process is provided in Figure~\ref{fig:overview}.
\vspace{-8pt}
\subsection{Composing Extremely Efficient Hybrid Model}
\vspace{-8pt}
To compose our hybrid model, we combine two complementary components: Mamba2~\cite{dao2024transformers} and MLA~\cite{liu2024deepseek} blocks. Each of these blocks contributes differently to efficiency and performance:
\vspace{-10pt}
\begin{itemize}[leftmargin=12pt, itemsep=0.5pt]
    \item \textbf{Mamba2 blocks} are based on SSMs with zero KV cache usage, making them ideal for long-context or memory-constrained settings. However, they often underperform when used exclusively~\cite{waleffe2024empirical}.
    \item \textbf{MLA blocks} compress standard attention process to reduce KV cache requirements while maintaining high performance. Yet, excessive compression can lead to noticeable performance drops~\cite{li2025x}.
\end{itemize}

Our hybrid architecture interleaves Mamba2 and MLA layers to set a balance between minimal memory usage and strong predictive performance. The composition process consists of two key stages: (\textit{i}) we first construct a pure Mamba2 model and a pure MLA model by replacing all attention blocks in the base Transformer with Mamba2 and MLA blocks, respectively, each initialized from the original pre-trained weights; (\textit{ii}) we then apply intermediate layer distillation (ILD) to align the outputs of each layer in the Mamba2 and MLA models with those of the corresponding layers in the base model, thereby preserving the original model’s knowledge during the transition.

Following this refined initialization, we compose the final hybrid model using our \textit{sensitivity measure-aware replacement of transformer} layers {(SMART)} layer selection strategy, which selects the optimal placement configurations of Mamba2 and MLA layers based on layer-wise sensitivity analysis. 


\vspace{-8pt}
\subsubsection{Refined Initialization} \label{Initialization}
\vspace{-8pt}

To initialize the MLA and Mamba2 blocks in pure MLA and pure Mamba2 models derived from a pre-trained base model, we introduce an enhanced initialization strategy that goes beyond conventional weight transfer. 


\vspace{-5pt}
\paragraph{Background: Multi-Head Attention }
Given an input hidden representation $H \in \mathbb{R}^{l \times d}$, MHA projects it into query, key, and value matrices using learned weights:
\begin{equation}
    Q = H W^Q, \quad K = H W^K, \quad V = H W^V,
\end{equation}
where $W^Q, W^K, W^V \in \mathbb{R}^{d \times n_h d_h}$ and $n_h$, $d_h$ denote the number of heads and per-head dimension, $d$ is the hidden dimension, and $l$ represents the sequence length. The attention operation computes:
\begin{equation}
    A = \text{Softmax}\left(Q K^\top/{\sqrt{d}}\right), \quad O = A V W^O,
\end{equation}
with $W^O \in \mathbb{R}^{d \times d}$. During inference, MHA requires caching $K$ and $V$ for all past tokens, incurring a memory cost of $2 n_h d_h l$.
\vspace{-5pt}
\paragraph{Background: Multi-Head Latent Attention}
MLA~\cite{liu2024deepseek} introduces a low-rank compression scheme to reduce memory usage. Instead of storing $K$ and $V$ directly, they are compressed into a latent vector:
\begin{equation}
    C^{KV} = H W^{DKV},
\end{equation}
with $W^{DKV} \in \mathbb{R}^{d \times r_{kv}}$ and $r_{kv} \ll n_h d_h$. Keys and values are reconstructed via:
\begin{equation}
    K^C = C^{KV} W^{UK}, \quad V^C = C^{KV} W^{UV},
\end{equation}
where $W^{UK}$ and $W^{UV}$ are up-projection matrices.
MLA can also compress the query:
\begin{equation}
    C^Q = H W^{DQ}, \quad Q^C = C^Q W^{UQ},
\end{equation}
where $C^Q \in \mathbb{R}^{l \times r_q}$ and $r_q \ll d$.
To retain compatibility with RoPE, MLA decouples positional encoding using separate projections:
\begin{equation}
    Q^R = \text{RoPE}(C^Q W^{QR}), \quad K^R = \text{RoPE}(H W^{KR}),
\end{equation}
where $W^{QR} \in \mathbb{R}^{r_q \times n_h d_r}$ and $W^{KR} \in \mathbb{R}^{d \times d_r}$ . The final queries and keys are then constructed as: $Q = [Q^C; Q^R], \quad K = [K^C; \text{repeat}(K^R)]$,
where $\text{repeat}(K^R)$ denotes replication across heads.
Overall, MLA reduces KV-cache size for inference from $O(n_h d_h l)$ to $O((r_{kv} + d_r) l)$. 
\vspace{-8pt}
\paragraph{Initializing MLA from Pre-trained MHA}\label{subsec:initialization}
To construct an MLA-based hybrid model from a pre-trained Transformer, we upcycle its attention modules by reparameterizing them into low-rank latent attention. This conversion is initialized using a structured approach based on singular value decomposition (SVD), enabling knowledge transfer while minimizing performance loss~\cite{li2025x}. The core MLA weights (i.e. query, key, and value projections) are initialized using low-rank approximations derived from pre-trained MHA parameters. For query weights, we apply SVD on $W^Q$:
\begin{equation}
    W^Q = U_q \Sigma_q V_q^\top,
\end{equation}
and set $W^{DQ} = U_q$. The up-projection matrices $W^{UQ}$ and $W^{QR}$ are derived from $\Sigma_q V_q^\top$ by partitioning and reshaping into the appropriate query and RoPE dimensions.
For key and value projections, we concatenate $W^K$ and $W^V$ and apply joint SVD:
\begin{equation}
    [W^K, W^V] = U_{kv} \Sigma_{kv} V_{kv}^\top.
\end{equation}
We then set $W^{DKV} = U_{kv}$, and derive $W^{UK}$ and $W^{UV}$ by partitioning and reshaping $\Sigma_{kv} V_{kv}^\top$. The shared RoPE key projection $W^{KR}$ is initialized from the average $W^K$ across heads. We choose a constant rank $r_q$, $r_{kv}$ across all layers.
Non-attention parameters such as the feedforward network and output projection $W^O$ are directly copied from the pre-trained model. Additional details of MLA initialization are provided in Appendix \ref{app:mla_init}.


\paragraph{Initializing Mamba2 from Attention Representations}

It has been shown in~\cite{wang2025mamballamadistillingaccelerating} that linear attention can be reinterpreted as a special case of SSMs. This connection enables the initialization of Mamba2 blocks from pre-trained attention-based Transformers.
In particular, the linearized form of attention without softmax resembles a linear RNN update:
\vspace{-5pt}
\begin{equation}
    h_t = m_{t-1, t} h_{t-1} + K_t^\top V_t, \quad y_t = \frac{1}{\sqrt{D}} Q_t h_t,
\end{equation}
which parallels the linear RNN structure:
\begin{equation}
    h_t = A_t h_{t-1} + B_t x_t, \quad y_t = C_t h_t.
\end{equation}
\vspace{-1pt}
To bridge this connection, MambaInLLaMA~\cite{wang2025mamballamadistillingaccelerating} proposes to initialize the continuous-time Mamba2 SSM parameters ($A$, $B$, $C$) from the weights of the attention blocks. This includes discretizing the SSM over a learned sequence of sampling intervals $\Delta_t$ to match the temporal dynamics of attention. Specifically, $K_t^\top V_t$ in attention is mapped to $B_t x_t$ in Mamba2; $Q_t$ in attention plays the role of $C_t$; and the memory coefficient $m_{t-1,t}$ corresponds to $A_t$ in the recurrent update. Additional details of MLA initialization are provided in Appendix \ref{app:mamba_init}. 


\vspace{-5pt}
\paragraph{Refined Initialization through Intermediate Layer Distillation (ILD)}
After initializing the parameters of MLA and Mamba2 layers from the pre-trained Transformer, we further refine their weights through a lightweight ILD procedure on a small portion of the training data, akin to the second phase of MOHAWK \cite{bick2024transformers}. This focused training aligns the internal representations between the MLA and Mamba2 layers and the pre-trained Transformer layers, ensuring a smoother start for downstream optimization and better preservation of the knowledge embedded in the original model. 
To perform ILD, we minimize the mean squared error (MSE) between the outputs of each pre-trained Transformer attention layer \( h^{\text{Attn}}_l \) and the corresponding outputs of MLA (\( h^{\text{MLA}}_l \)) and Mamba2 (\( h^{\text{M2}}_l \)) layers. The ILD losses for Mamba2 and MLA are defined as:
\begin{equation}
\small
\begin{split}
 \mathcal{L}^{\text{\small{Mamba2}}}_{\text{ILD}} = \sum_{l\in\{1,2,...,L\}} \left\| h^{\text{Attn}}_l - h^{\text{M2}}_l \right\|_2^2 ,~~
 \mathcal{L}^{\text{\small{MLA}}}_{\text{ILD}} = \sum_{l\in\{1,2,...,L\}} \left\| h^{\text{Attn}}_l - h^{\text{MLA}}_l \right\|_2^2,
\end{split}
\end{equation}
where $L$ denotes the total number of layers and $h_l$ refers to the output of the $l^{th}$ layer. Unlike MOHAWK, where the weights of MLP within each layer remain frozen during distillation, we allow training of all parameters in the Mamba2 and MLA layers. 
The refined initialization resulting from this ILD has proven to be crucial for enhancing the subsequent end-to-end distillation process, as evidenced in Section~\ref{ablation:ILD}.



\vspace{-5pt}
\subsubsection{\smart: Sensitivity Measure-Aware Replacement of Transformer Layers}

\begin{wrapfigure}{R}{0.45\textwidth}
    \centering
    \vspace{-0.2in}
    \includegraphics[width=0.45\textwidth]{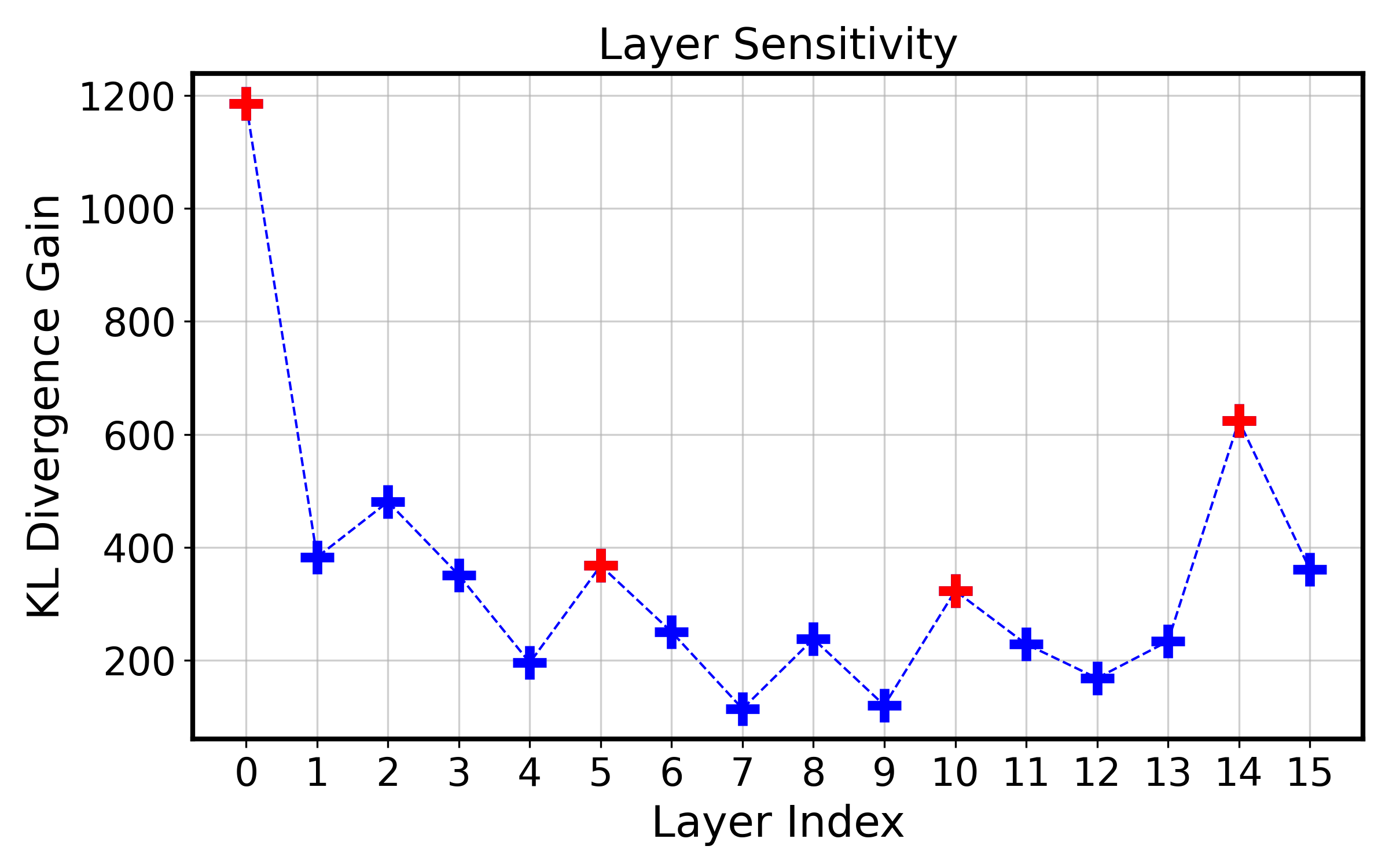}
    \vspace{-0.3in}
    \caption{\small{Layer sensitivity scores for Llama3.2-1B using 4096 samples from the validation dataset. Red markers indicate the MLA layer indices selected by our \smart strategy with $N=4$.}}
    \vspace{-10pt}
  \label{fig:sensitivity}
\end{wrapfigure}
The final stage of our initialization process is to compose our hybrid model from the full MLA and full Mamba2 models.
Let $N$ denote the desired total number of MLA layers. The set of layers assigned as MLA is represented by $\mathcal{L}_{\text{MLA}} \subseteq \{0, 1, \ldots, L{-}1\}$. To minimize the performance gap between the hybrid and fully attention-based models, we introduce a Sensitivity Measure-Aware Replacement of Transformer Layers (\smart) strategy, which leverages empirical sensitivity analysis to guide the layer assignment. 
To measure the sensitivity of each layer, we begin with the full Mamba2 model after ILD. First, we compute the KL divergence between the full attention-based teacher model and the student model where all layers are Mamba2. Then, for each layer $i$, we construct a variant of the student model in which only the $i^\text{th}$ layer is replaced with MLA while the rest remain Mamba2, and measure the KL divergence against the teacher. The sensitivity gain $s_i$ for layer $i$ is defined as the reduction in KL divergence relative to the full Mamba2 model. 
The sensitivity score is formally computed as:
\begin{equation}
    \begin{split}
       s_i = \sum_{t=1}^T \mathbf{KL}\left[p(\cdot \mid y_{1:t}, x, \theta_T) \,\middle\|\, p(\cdot \mid y_{1:t}, x, \theta)\right] - 
        \mathbf{KL}\left[p(\cdot \mid y_{1:t}, x, \theta_T) \,\middle\|\, p(\cdot \mid y_{1:t}, x, \theta'_i)\right],
    \end{split}
\end{equation}
where $\theta_T$ and $\theta$ denote the parameters of the teacher and full Mamba2 model, and $\theta'_i$ corresponds to the variant with MLA inserted at layer $i$. In this equation, \( y_{1:t} \) refers to model predictions up to time step \( t \) and \( x \) is the input sequence and \( T \) is total number of decoding time steps.
A higher score indicates that the $i^\text{th}$ layer plays a more critical role in aligning with the teacher and should thus be prioritized for MLA replacement due to its higher representational capacity.
Figure~\ref{fig:sensitivity} shows an example sensitivity profile for the \ours 1B model after refined initialization, where we observe that the earliest and latest layers tend to exhibit higher sensitivity, while middle layers are typically less critical. 
While it may seem intuitive to simply select the top $N$ layers with the highest sensitivity scores, this strategy often leads to suboptimal layer placements—especially if it results in large gaps between MLA layers. To enforce a more structured distribution and preserve the most sensitive positions, we adopt the following three-step procedure:
\vspace{-5pt}
\begin{itemize}[leftmargin=10pt, itemsep=1pt]
    \item \textbf{Terminal Preservation:} 
    We divide the total \( L \) layers into \( N \) roughly uniform partitions. We take the first and last $\lfloor L/N \rfloor$ layers in the first and last partitions. From these two partitions, we select the layer with the highest sensitivity score as the first and last MLA layers, denoted \( L_1^{\text{MLA}} \) and \( L_N^{\text{MLA}} \), respectively. This ensures that the most sensitive edge layers are preserved.
    \item \textbf{Near-Uniform Intermediate Distribution:} 
    Given the range \( [L_1^{\text{MLA}} + 1, L_N^{\text{MLA}} - 1] \), we aim to place the remaining \( N - 2 \) MLA layers such that the gaps between consecutive MLA layers are as uniform as possible. We constrain the gap between adjacent MLA layers to lie within the range of min: $\lfloor \frac{L_N^{\text{MLA}}-L_1^{\text{MLA}}-N+1}{N-1} \rfloor$, and max: $\lceil \frac{L_N^{\text{MLA}}-L_1^{\text{MLA}}-N+1}{N-1} \rceil$. We enumerate all valid configurations \( \{C_j\}_{j=1}^{M} \) that meet this spacing constraint, where each candidate \( C_j \) defines a possible set of intermediate MLA layer indices.    
    \item \textbf{Maximal Sensitivity Scores:} For each valid configuration, we compute the total sensitivity score and choose the one with the highest cumulative score: $C^* = \underset{C_j}{\arg\max} \sum_{i\in C_j} s_i$. 
\end{itemize} 
\vspace{-10pt}
Additional details and examples of this layer selection procedure are provided in Appendix~\ref{apndx:SMART}.
\vspace{-2pt}
\subsection{Efficient Training}

After initialization and model composition, we follow an end-to-end knowledge distillation and DPO training stages to incrementally improve model accuracy and efficiency.  
\vspace{-10pt}
\paragraph{End-to-End Knowledge Distillation}  This stage involves an end-to-end distillation with supervised fine-tuning (SFT): 
\vspace{-10pt}
\begin{equation}
    \mathcal{L}_{\theta} = \sum_{t=1}^T \mathbf{KL}[p(\cdot | y_{1:t}, x, \theta_T) || p(\cdot | y_{1:t}, x, \theta) ], 
\end{equation}
where $\theta$ and $\theta_T$ are the parameters of the student model and the teacher model respectively. Such distillation step is crucial for transferring the rich, pre-trained knowledge from the teacher model. 
\vspace{-10pt}
\paragraph{Direct Preference Optimization (DPO)} In the final training stage, we perform DPO which is a binary cross entropy loss to adjust the preference of the student model. We set the distilled student model itself as the reference model as it makes the training much stabler.

\vspace{-3pt}
\section{Experiments and Results}
\vspace{-3pt}
\subsection{Training Setup}
\vspace{-5pt}
All of our \ours models are distilled from the Llama family: Llama3.2-1B-Instruct, Llama3.2-3B-Instruct, and Llama3.1-8B-Instruct for our experiments. For ILD and SFT, we use the same dataset as in \cite{wang2025mamballamadistillingaccelerating} which includes multiple public datasets such as \textit{OpenHermes-2.5}\cite{OpenHermes2_5}, \textit{GenQA}\cite{chen2024genqa}, and \textit{Infinity-Instruct} \cite{infinity_struct}, with a total number of 6.8 billion tokens. The dataset is splited into $20\%$ and $80\%$ for ILD and SFT separately. We repeat the same training data more than one epoch to match the desired token budget. For DPO preference tuning, we adopt three datasets \textit{Llama3-ultrafeedback}\cite{ultrafeedback_armorm}, \textit{orca\_dpo\_pairs}\cite{OpenOrca}, and \textit{ultrafeedback\_binarized}\cite{cui2023ultrafeedback}. All models were trained on a single node equipped with eight AMD MI300 GPUs. Our training details are provided in Appendix~\ref{app:hyper_param}.

\vspace{-5pt}
\subsection{Performance Evaluation}
\vspace{-5pt}
\paragraph{Evaluation Tasks} 
We adopt the LM Harness Eval benchmark ~\cite{gao2023} to perform zero-shot and few-shot evaluations on language understanding tasks, which includes ARC-Challenge (ARC)~\cite{clark2018think}, ARC-Easy (ARE)~\cite{clark2018think},HellaSwag (HS)~\cite{zellers2019hellaswag}, MMLU (MM)~\cite{hendrycks2020measuring}, OpenBookQA (OB) ~\cite{mihaylov2018can}, PIQA~\cite{bisk2020piqa}, RACE (RA)~\cite{lai2017race}, and WinoGrande (WG) ~\cite{sakaguchi2021winogrande}. We provide more details for model evaluations in Appendix~\ref{app:eval_details}. 

\begin{table*}[t]
     \setlength\extrarowheight{2pt}
    \centering
    \scalebox{0.63}{
    \begin{tabular}{l|cccc|ccccccccccc}
        \toprule
        Model and Setting & Teacher & Tokens & Size & KV Size & ARC & ARE & HS & MM & OB & PI & RA & WG & Avg. \\        \midrule
        \midrule
        Llama3.2-1B-Inst & - & 9T & 1.24B & 100\% &37.97 &	63.30 & 60.65 & 46.05 & 34.80 & 74.32 & 38.18 & 59.67 & \textbf{51.87} \\
        \hdashline
        MambaInLlama-1B-50\%* & 8B  & 7B & 1.27B & 50\% & 40.78	& 65.57 &	61.4 &	40.17 &	39.2 &	74.32 &	38.28 &	58.72 &	52.31 \\
        X-EcoMLA-1B ($r_{kv}=64$) & 8B  & 7B & 1.23B & 9.37\% &  40.02  &	67.17 &	58.4 & 38.53 &	37.8 &	73.83 &	39.43 &	60.93 &	52.01 \\
        Llamba-1B & 1B+70B & 8B & 1.41B & 0\% & 37.12&	65.36&	61.31&	38.11&	36.8&	73.78&	37.61&	60.62&	51.34 \\ 
        \hdashline
        \rowcolor{gray!20}
        \ours-1B, 8MLA-8M2 & 8B & 7B & 1.27B & 7.81\% & 42.49&	67.38&	60.54&	38.94&	41.6&	72.91 &	38.37&	61.25&	52.94 \\ 
        \rowcolor{gray!20}
        \ours-1B, 6MLA-10M2  & 8B & 7B & 1.28B & 5.86\% &  43.94&	67.51&	60.46&	38.21&	41.2&	73.23&	37.61&	61.17&	52.92 \\
        \rowcolor{gray!20}
        \ours-1B, 4MLA-12M2 & 8B & 7B &  1.28B & 3.91\% & 42.32 &	66.96 &	58.93&	37.91&	40.6&	72.96&	37.7&	58.88&	52.03 \\
        \bottomrule
        \toprule
        Llama3.2-3B-Inst & - & 9T & 3.21B & 100\% & 46.08&	67.93&	70.38&	60.34&	36.4&	75.79&	40.86&	67.25&	\textbf{58.13} \\
        \hdashline
        MambaInLlama-3B-50\% & 70B  & 20B & 3.35B & 50\% & 54.1&	80.26&	74.45&	52.47&	42.4&	77.69&	43.44&	67.32&	61.52 \\
        X-EcoMLA-3B ($r_{kv}=816$) & 3B  & 7B & 3.21B & 42.97\% &  48.38&	70.37&	72.41&	57.51&	38.2&	76.28&	46.41&	68.11&	59.71 \\
        X-EcoMLA-3B ($r_{kv}=128$)* & 8B  & 7B & 3.21B & 9.37\% &  52.05&	75.38&	70.95&	53.2&	40.8&	77.09&	44.69&	66.85&	60.13 \\
        Llamba-3B & 3B+70B & 10B & 3.66B & 0\% & 45.65&	73.78&	73.31&	52.32&	42.4&	78.02&	40.1&	70.01&	59.45 \\ 
        \hdashline
        \rowcolor{gray!20}
        \ours-3B, 14MLA-14M2 & 8B & 9B & 3.27B & 4.69\% & 51.28&	76.14&	72.57&	52.1&	42.4&	77.53&	45.93&	67.56& 60.69 \\ 
        \rowcolor{gray!20}
        \ours-3B, 8MLA-20M2  & 8B & 9B & 3.36B & 2.68\% &  51.96&	75.97&	72.38&	48.16&	42.8&	77.64&	43.54&	65.67&	59.77\\ 
        \rowcolor{gray!20}
        \ours-3B, 6MLA-22M2 & 8B & 9B &  3.39B & 2.01\% & 50.77 &	76.09&	71.46&	50.06&	43.4&	77.26&	42.49&	66.46&	59.75\\
        \bottomrule
        \toprule
        Llama3.1-8B-Inst & - & 15T & 8.03B & 100\% & 54.86&	79.55&	79.23&	68.13&	43&	80.9&	44.69&	73.88&	\textbf{65.53} \\
        \hdashline
        MambaInLlama-8B-50\% & 70B  & 20B & 8.3B & 50\% & 59.73&	84.81&	79.69&	59.74&	44&	80.03&	46.12&	74.11&	66.03\\
        Minitron-8B & 15B$^\dagger$  & 94B & 8.3B & 66.67\% & 52.73 &	79.5 &	77.4 &	62.95 &	45.2 &	81.39 &	39.71 &	72.69 & 63.95\\
        X-EcoMLA-8B ($r_{kv}=128$)* & 8B  & 7B & 8.03B & 9.37\% &  56.57&	79.04&	77.38&	58.6&	42.8&	79.6&	48.33&	70.96&	64.16\\
        Llamba-8B & 8B+70B & 12B & 8.32B & 0\% & 53.84 &79.71&	76.25&	60.29&	44&	79.16&	40.38&	72.77&	63.3 \\ 
        \hdashline
        \rowcolor{gray!20}
        \ours-8B, 16MLA-16M2 & 8B & 11B & 8.19B & 5.47\% & 58.62&	78.37&	79.27&	58.17&	43.4&	80.03&	49.28&	72.61&	64.97\\ 
        \rowcolor{gray!20}
        \ours-8B, 8MLA-24M2  & 8B & 11B & 8.38B & 2.73\% &  58.87&	79.17&	78.4&	54.6&	43.6&	79.43&	46.22&	72.45 &	64.12\\ 
        \bottomrule

    \end{tabular}
    }
    \vspace{-6pt}
    \caption{{\small{Zero-shot evaluation on the LM Harness Eval benchmark across eight tasks: ARC-Challenge (ARC), ARC-Easy (ARE), HellaSwag (HS), MMLU, OpenBookQA (OB), PI, RACE (RA), and WinoGrande (WG). All the teacher models are Llama3.1-8B/70B or Llama3.2-1B/3B execpt for Minitron (which is $^\dagger$Nemotron-15B)}. $^*$The X-EcoMLA results are reproduced by ourselves. }}
    \label{tab:main_results}
    \vspace{-15pt}
\end{table*}

\vspace{-5pt}
\paragraph{Zero-shot Results}
The results of our zero-shot evaluations are summarized in Table~\ref{tab:main_results}. We compare our \ours with the base Llama models and other baselines based on distillation: MambaInLLaMA (Hybrid Mamba2-GQA)\cite{wang2025mamballamadistillingaccelerating}, X-EcoMLA (Pure MLA)\cite{li2025x}, Llamba (Pure Mamba2) \cite{bick2025llamba}, and Minitron (Pruning) \cite{muralidharan2024compact}. Besides the evaluation results, we list the teacher model size, number of training tokens, student model size (the number of parameters), and the KV cache size compared to the base Llama models. For the MLA layers in \ours, we set $r_{kv}=128$ for the 1B and 3B models and $r_{kv}=160$ for the 8B models. For each model size, we tested various combinations of MLA and Mamba2 layers. 

As shown in Table~\ref{tab:main_results}, compared to the base Llama models, our \ours  achieves extreme KV cache compression without noticeable performance drop. For 1B and 3B models, we achieves $3.91\%$ ($25.6\times$ compression) and $2.01\%$ ($49.78\times$ compression) KV cache size with even higher performance than the base Llama models. For the 8B models, we reach $5.47\%$ ($18.3\times$ compression) and $2.73\%$ ($36.6\times$ compression) KV cache size with only {$1\%$ and $2.15\%$} performance drop. Note that for the 8B models, we achieve such compression ratio using only same size teachers. 
Further more, our \ours achieves a better balance between KV cache compression and performance with much more efficient training (i.e., fewer training tokens or smaller teacher size) than other distillation-based methods. For example, compared with \textit{hybrid MambaInLlama}, \ours has similar performance with  $12.79\times$ and $24.88\times$ smaller KV cache for 1B and 3B models. For 8B models, \ours achieves $9.14\times$ KV cache compression with only $1.6\%$ performance degradation by using a much smaller teacher (8B) and fewer training tokens (11B). Similarly, our method has roughly the same performance as the pure-MLA X-EcoMLA models, but with up to $3.4\times$ less KV cache size. Moreover, compared to the pure Mamba model Llamba-8B, \ours shows significantly better performance with both smaller teacher and fewer training tokens with minimal KV cache overhead. 
Moreover, we further compare \ours with state-of-the-art hybrid models trained from scratch in Appendix~\ref{sec:main_results_pretrain} (see Table~\ref{tab:main_results_pretrain}). While prior methods like SAMBA and Mamba-2-Hybrid rely on 1.5T–3.5T training tokens, our \ours models achieve competitive or superior performance using only 7–11B tokens—representing a \textbf{214×–500× reduction} in training data. 

\vspace{-8pt}
\begin{wraptable}{R}{0.55\textwidth}
    \centering
    \vspace{-0.1in} 
    
    \scalebox{0.6}{
    \begin{tabular}{l |c|ccccc|c}
        \toprule
        Model and Setting   & KV \% & ARC & HS& MM & WG & TQ  & Avg. \\  
        \midrule
        \midrule
        Llama3.1-8B-Inst   & 100\% &60.75 &	80.12 &	68.23 &	73.72 &	53.99 	 &	67.36   \\
        \hdashline
        Minitron  & 66.7\% &49.49 &	81.61 &	64.34 &	72.77 &	43.97 &  62.44 \\ 
        MambaInLlama-50\%   & 50\% & 60.41 &	77.97 &	56.67 &	71.35 &	66.6 &  66.6\\
        MambaInLlama-25\%   & 25\% &59.22 &	76.88 &	53.94 &	64.88 &	64.64  & 63.91 \\
        MambaInLlama-12.5\%  & 12.5\% & 53.33 &	72.16 &	50.85 &	63.61 &	61.12  & 60.21 \\
        MambaInLlama-0\%   & 0\% & 53.51 & 70.31 & 44.21 & 58.91 & 52.31 &   55.85 \\
        Llamba  & 0\% & 57.94 &	77.13 &	59.89 &	72.77 &	49.46 &	 63.44 \\ 
        X-EcoMLA-8B ($128$)*   & 9.37\% & 59.64	& 76.9 &	58.73&	71.43&	60.86& 	  65.51\\
        \hdashline
        \rowcolor{gray!20}
        \ours, (16-16)     &  5.47\% &60.49 &	78.29 &	58.84 &	71.98 &	64.28 &   66.78\\ 
        \rowcolor{gray!20}
        \ours, (8-24)     & 2.73\% & 60.41 &	76.11 &	55.06 &	71.11 &	61.01 &   64.74	 \\ 
        \bottomrule
    \end{tabular}
    }
    \caption{\small{Few-shot evaluation on the LM Harness Eval benchmark across five tasks.}}
    \label{tab:main_results_fewshot}
    
    \vspace{-3mm} 
\end{wraptable}
\paragraph{Few-shot Results}
In Table~\ref{tab:main_results_fewshot}, we report the results of few-shot evaluations on the \ours-8B models under 25-shot ARC-Challenge (ARC), 10-shot HellaSwag (HS), 5-shot Winogrande (WG),  5-shot MMLU (MM), and 0-shot mc2 for Truthful-QA (TQ) tasks. Among all models, our \ours achieves the best performance with only $5.47\%$ KV cache usage. The closest one to us is MambaInLlama-50\% but it's trained with $1.8\times$ more tokens and has $9.14\times$ more KV cache usage. For pure Mamba2 models such as MambaInLlama-8B-0\% and Llamba-8B, they don't use any KV cache but their performance is significantly worse than \ours even with more training tokens and larger teachers. {Complete few-shot results for \ours-1B and \ours-3B can be found in Appendix \ref{appendix:fewshot}.}

\vspace{-5pt}
\paragraph{{Results on GSM8k}}
{In Table~\ref{table:gsk_eval}, we evaluate the math-reasoning ability of \ours via 8-shot GSM8K task, comparing its performance against baseline models. For the Zebra-Llama-1B architecture, the 8MLA-8Mamba2 configuration achieves the highest accuracy, even outperforming the base Llama-3.2-1B-Instruct model by 6.7\%. Increasing the number of MLA layers to 12 results in a performance drop from $41.09\%$ to $29.57\%$; however, this result remains superior to both MambaInLlama-1B and Llamba-1B by $4.63\%$ and $6.75\%$, respectively. For Zebra-Llama-3B, it is slightly worse than X-EcoMLA-3B given more than 2x smaller KV cache. However, our Zebra-Llama is still 9.71\% and 15.01\% better than MambaInLlama-3B and Llamba-3B with 14 MLA layers and 14 Mamba2 layers. Even we decrease the number of MLA layers to 8, which leads to only 2\% of the original KV cache, our Zebra-Llama is still 2.1\% and 7.43\% better than MambaInLlama-3B and Llamba-3B. This consistent superiority demonstrates that our hybrid model structure and tailored training strategy effectively preserve, and in many cases enhance, the math-reasoning capabilities of the underlying base model while maintaining high KV cache efficiency. }

\begin{figure}[t]
  \centering
  \begin{minipage}[t]{0.47\textwidth}
    \centering
        \includegraphics[height=1.5in]{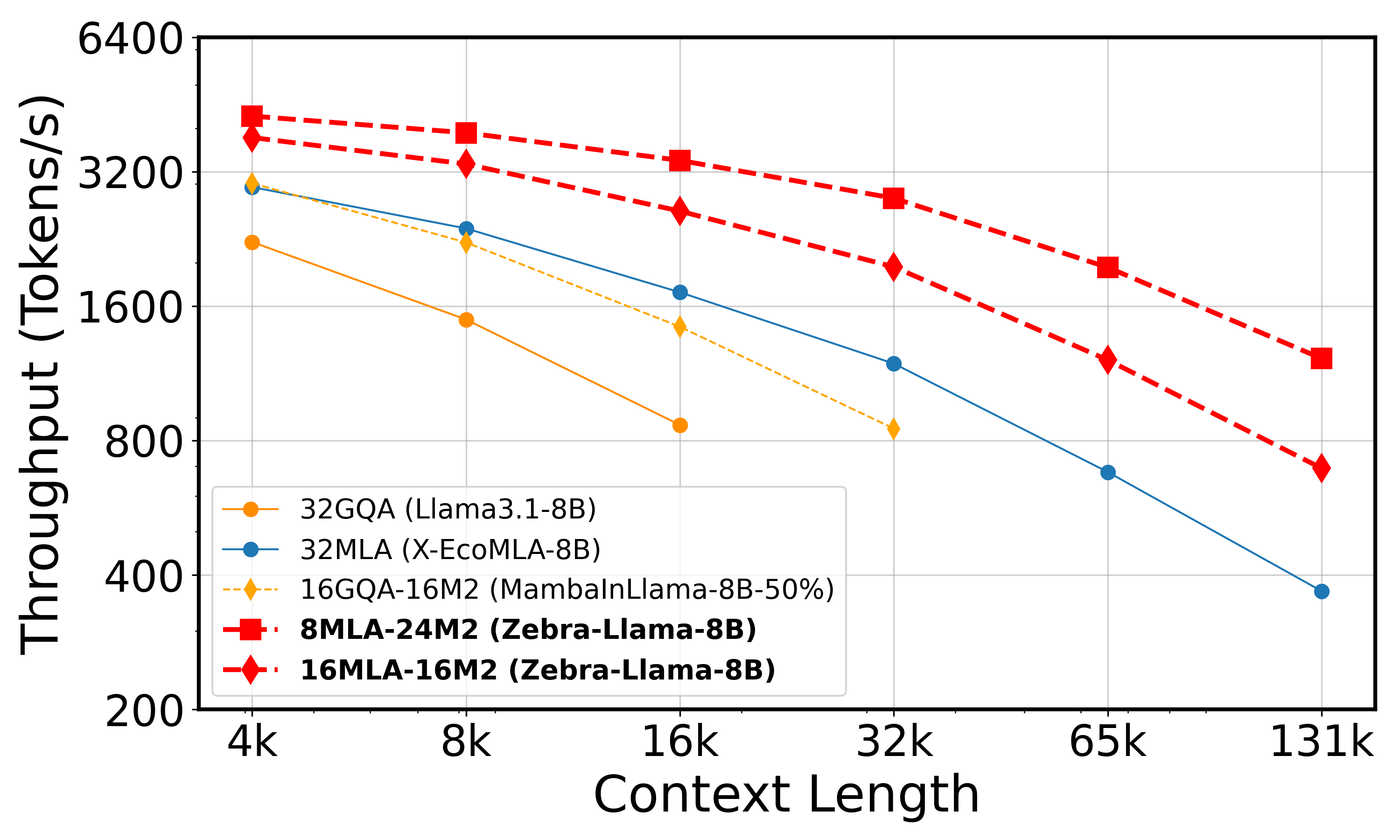}\vspace{-2mm}
        \caption{\small{Inference throughput vs. {context} length of various 8B-size models. We measure the throughput under batch size {48 and output length 1024}.}}\label{fig:infer}
        \vspace{-5mm}
  \end{minipage}
  \hfill
  \begin{minipage}[t]{0.47\textwidth}
    \centering
        \includegraphics[height=1.5in]{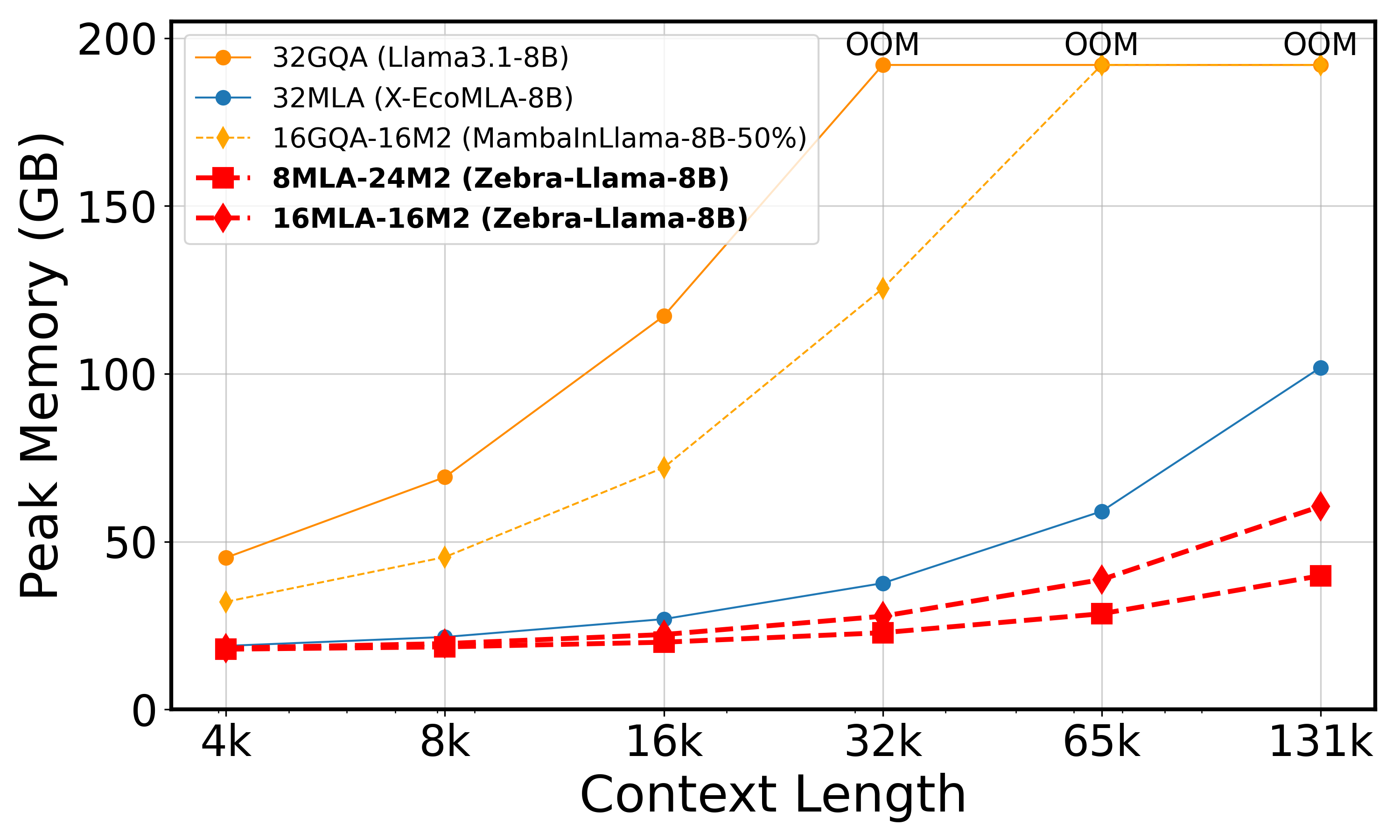}\vspace{-2mm}
    \caption{\small{Inference peak memory vs. {context} length of various 8B-size models. The out-of-memory scenarios are marked with 'OOM'.}} \label{fig:infer_memory} 
        \vspace{-5mm}
  \end{minipage}
  \vspace{-5pt}
\end{figure}

\vspace{-8pt}
\paragraph{{Throughput Evaluation}}

\begin{wraptable}{R}{0.42\textwidth}
    \centering
    \vspace{-0.1in} 
    
    \scalebox{0.65}{
    \begin{tabular}{c|cc|c}
        \toprule
        Model & KV \% & Tokens & GSM8K (8) \\ 
        \midrule \midrule
        Llama-3.2-1B-Instruct &100\% & 9T & 38.51\\
        MambaInLlama-1B &50\% & 7B & 24.94\\ 
        X-EcoMLA-1B &9.37\% & 7B & 38.06\\
        Llamba-1B &0\% & 8B & 22.82\\
        \rowcolor{gray!20}
        Zebra-Llama-1B (8-8) &7.81\% & 7B & 41.09\\
        \rowcolor{gray!20}
        Zebra-Llama-1B (4-12) &3.91\% & 7B & 29.57\\
        \hdashline
        Llama-3.2-3B-Instruct &100\% & 9T &70.89\\
        MambaInLlama-3B &50\% & 20B &53.06\\ 
        X-EcoMLA-3B &9.37\% & 7B & 65.28\\
        Llamba-3B &0\% & 10B & 47.76\\
        \rowcolor{gray!20}
        Zebra-Llama-3B (14-14) &4.69\% &  9B & 62.77 \\
        \rowcolor{gray!20}
        Zebra-Llama-3B (6-22) &2.01\% & 9B & 55.19\\
        \bottomrule
    \end{tabular}
    }
    \caption{\small{{8-shot GSK8K results for \ours and baselines.}}}
    \label{table:gsk_eval}
    \vspace{-3mm} 
\end{wraptable}
Figures \ref{fig:infer} and \ref{fig:infer_memory} detail the inference efficiency of \ours in terms of throughput and memory consumption for the models that are distilled from Llama-3.1-8B. All experiments are conducted on a single AMD MI300X GPU with 192GB memory. We fix the batch size to 48 and the output length to 1024 and measure the inference throughput across various context/input lengths. As shown, our \ours significantly outperforms the base Llama-3.1-8B model by 3.9$\times$ for throughput with only 17\% peak memory at 16k input tokens. Similarly, \ours outperforms the MambaInLlama model by $3.28\times$ in throughput with only $18.2\%$ of its peak memory at 32k input tokens. Beyond 16k and 32k tokens, the Llama-3.2-8B model and MambaInLlama-8B model will both run out of memory. In constrast, our Zebra-Llama model only occupied 62.68GB (8MLA-24Mamba2) and 104.6GB (16MLA-16Mamba2) even with 131k input tokens, showcasing its brilliant ability for memory efficiency. 


\vspace{-8pt}
\subsection{Ablation Studies}
\vspace{-6pt}
In this section, we present a series of ablation studies aimed at justifying key design decisions in our approach. Specifically, we examine the impact of initialization strategies, the effectiveness of our \smart layer selection mechanism, the trade-offs between the number and size of MLA and Mamba2 layers, and the role of teacher model scaling. 


\vspace{-7pt}
\subsubsection{Effect of Initialization Strategies}
\vspace{-5pt}

\begin{wraptable}{R}{0.42\textwidth}
    \centering
    \vspace{-0.4in} 
    \scalebox{0.6}{
    \begin{tabular}{c|c|c|c}
        \toprule
        Selection Strategy & MLA Indices & Total sen. & Avg. Score \\    \midrule \midrule
        Uniform \#1 &{[}0,4,8,12{]} & 1787.9 & 48.84 \\
        Uniform \#2 &{[}3,7,11,15{]} & 1055.8 & 49.72 \\ 
        Max score &{[}0,1,2,14{]} & 2672.5 & 48.98 \\
        \hdashline
        Possible middle \#1&{[}0,4,9,14{]}& 2125.7 &  49.76 \\
        Possible middle \#2&{[}0,5,9,14{]} & 2297.5 &  49.95 \\
        \rowcolor{gray!20}
        SMART (ours) &{[}0,5,10,14{]} & 2500.2 &	\textbf{50.15} \\
        \bottomrule
    \end{tabular}
    }
    \caption{\small{SFT results with different layer selection strategies on Llama-3.2-1B model. Except for the selected MLA indices, all models are trained with same configuration.}}
    \label{ablation:layer_selection}
    \vspace{-5mm} 
\end{wraptable}

\begin{figure}[t]
  \centering
  \begin{minipage}[t]{0.47\textwidth}
    \centering
        \includegraphics[height=1.5in]{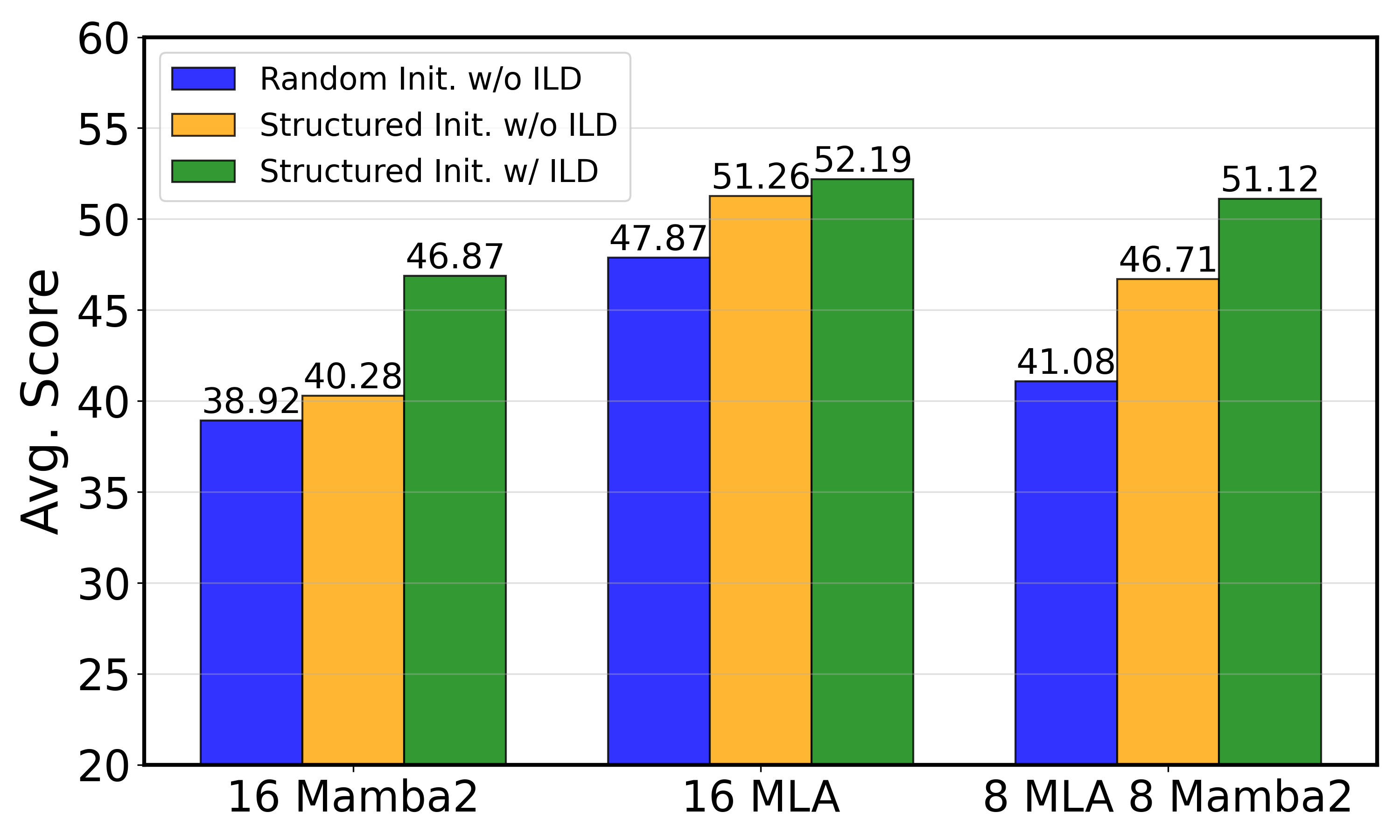}\vspace{-2mm}
    \caption{\small{Performance of various initialization strategies after SFT for different model architectures. Generally, our proposed two-stage method achieves the highest average scores.}} \label{fig:init} 
        \vspace{-5mm}
  \end{minipage}
  \hfill
  \begin{minipage}[t]{0.47\textwidth}
    \centering
    \includegraphics[height=1.55in]{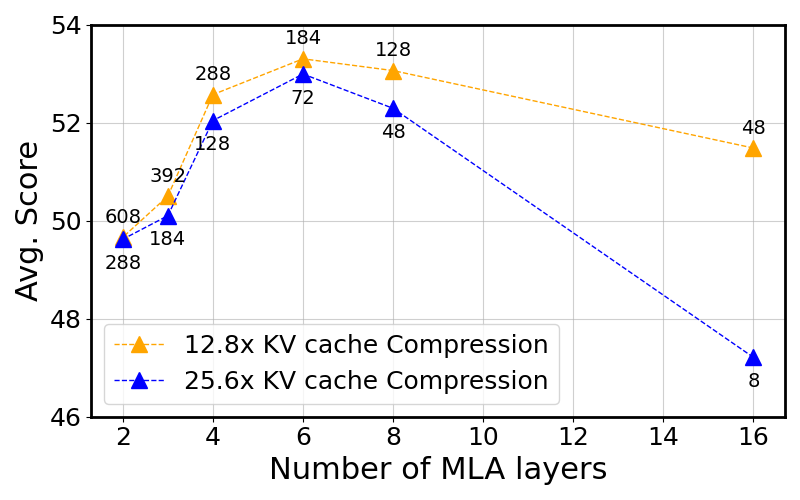}\vspace{-2mm}
    \caption{\small{Tradeoff between KV rank $r_{kv}$ and the number of MLA layers when fixing the total KV-cache size of given model.}}  \label{fig:tradeoff}
        \vspace{-5mm}
  \end{minipage}
\end{figure}

Figure \ref{fig:init} presents our assessment of various initialization methodologies through a comparison of three scenarios: {Random weight initialization without ILD}, {Structured weight initialization without ILD}, and the proposed {Structured weight initialization with ILD}. 

The results highlight that both structured weight initialization and ILD significantly boost SFT performance, especially when used together. For Mamba layers, ILD is crucial for aligning their outputs with the original models due to architectural differences from attention layers, providing the primary performance uplift. As for MLA layers, structured initialization can boost the accuracy significantly, which offers a strong foundation further refined by ILD.




\vspace{-4pt}
\subsubsection{Impact of \smart Layer Selection}
\vspace{-6pt}


In Table \ref{ablation:layer_selection}, we demonstrate the benefits of our \smart layer selection strategy by comparing with other layer selections for our \ours-1B with 4 MLA layers, by using the configs shown in Figure \ref{fig:sensitivity} as an example. 
First, by comparing the first two layer selections sets and our optimal solutions, we can conclude that \textit{\textbf{Terminal Preservation}} strategy (i.e., always selecting layers from the first and last few layers) are important for preserving model accuracy. Second, we also testify the naive greed selection greedy method (the third row); the results shows that  uniformly distributing the MLA layers has a major contribution to final performance as well, which verified the effectiveness of our \textit{\textbf{Near-Uniform Intermediate Distribution }} strategy. Third, by analyzing the last three selections, together with {{Terminal Preservation}} and {{Near-Uniform Intermediate Distribution }}, \textbf{\textit{Maximal Sensitivity Scores}} is indeed a good indicator for final accuracy score.  In short, these outcomes underscore how each of the three pillars--when appropriately combined--contributes critically to the \smart layer selection strategy's effectiveness.
 {While we believe that \smart offers a systematic and principled alternative to heuristic layer selection, we look forward to further exploration and advancement of this direction in future work.}


\vspace{-6pt}
\subsubsection{Tradeoff between number of MLA layers and $r_{kv}$}
\vspace{-6pt}
The KV-cache size in \ours is determined by two factors: the number of MLA layers and the KV rank $r_{kv}$ of each MLA layer. 
Figure \ref{fig:tradeoff} presents our findings from varying these two factors for our 1B model while maintaining a constant total KV-cache size. We observed that for significant KV-cache compressions, such as $12.8\times$ and $25.6\times$, optimal performance typically occurs with a moderate number of MLA layers, around six, coupled with an intermediate $r_{kv}$. Deviating from this balance will hurt the model's performance. On one hand, increasing the number of MLA layers excessively, like to 16 layers at a $25.6\times$ compression, forces the $r_{kv}$  per layer to become very small (e.g., $r_{kv}$ =8), which significantly degrades performance. On the other hand, reducing the number of MLA layers too much also leads to a decline in performance,  since the model will then consist almost entirely of Mamba2 layers, which typically have a higher capacity gap with MHA layers than MLA layers do.

\vspace{-6pt}
\subsubsection{Scaling the Teacher}
\vspace{-6pt}
Effective knowledge transfer from a well-chosen, pre-trained teacher model is crucial for our method's success. As shown in Table~\ref{ablation:teacher}, appropriately scaled teachers significantly enhance student model performance though with a diminishing return when gradually increasing size of teacher models. 
{This is due to the "capacity gap" phenomenon in knowledge distillation, where a smaller student model struggles to fully absorb and generalize the complex teacher's representations when the teacher model far exceeds student mimicry capabilities~\cite{mirzadeh2020improved, jafari2021annealing}. 
Therefore, it's vital to select a teacher model that is sufficiently powerful to provide rich knowledge but not so extremely larger than the student model, thereby balancing distillation efficacy with student-teacher compatibility. Investigating adaptive teacher scaling or multi-stage distillation offers future solutions to these capacity limitations.}

\begin{table*}[t]
     \setlength\extrarowheight{2pt}
    \centering
    \scalebox{0.64}{
    \begin{tabular}{lc|ccccccccccc}
        \toprule
        Model and Setting & Teacher Size & ARC & ARE & HS & MM & OB & PI & RA & WG & Avg. \\        \midrule
        \midrule
        Llama3.2-1B-Inst & - &37.97 &	63.30 & 60.65 & 46.05 & 34.80 & 74.32 & 38.18 & 59.67 & \textbf{51.87} \\
        \hdashline
        \ours-1B, 4MLA-12M2 & 1B  & 38.91&	61.7& 55.03& 33.83 &	37.2&	71.93&	35.41&	58.88&	49.11 \\ 
        \ours-1B, 4MLA-12M2 & 3B & 39.51 &	62.79&	57.61&	37.94&	38.20&	72.52&	36.94&	56.59 & 50.26 \\ 
        \ours-1B, 4MLA-12M2 & 8B & 42.32 &	66.96 &	58.93&	37.91&	40.6&	72.96&	37.7&	58.88&	52.03 \\
        \ours-1B, 4MLA-12M2 & 70B & 43.17&	69.57&	57.77&	39.45&	38.80&	72.80&	38.09&	59.83 & 52.44\\
        \midrule
        Llama3.2-3B-Inst & - & 46.08&	67.93&	70.38&	60.34&	36.4&	75.79&	40.86&	67.25&	\textbf{58.13} \\
        \hdashline
        \ours-3B, 8MLA-20M2  & 3B  &  45.48&	69.28&	69.04&	47.69&	40.80&	74.81&	42.01&	63.38& 56.56 \\ 
        \ours-3B, 8MLA-20M2 & 8B & 51.54&	75.55&	71.52&	47.12&	43.6&	77.2&	42.68&	65.9&	59.39\\ 
        \ours-3B, 8MLA-20M2 & 70B &51.96&	77.23&	69.46&	48.32&	43.4&	76.01&	43.35&	65.19& 59.37 \\ 
        \midrule
        Llama3.1-8B-Inst & - & 54.86&	79.55&	79.23&	68.13&	43&	80.9&	44.69&	73.88&	\textbf{65.53} \\
        \hdashline
        \ours-8B, 8MLA-24M2  & 8B  &  56.48&	78.79&	76.84&	53.72&	44.4&	79.43&	44.31&	70.64 & 63.08\\ 
        \ours-8B, 8MLA-24M2 & 70B & 58.53&	80.72&	76.64&	53.82&	45.4&	80.03&	43.06&	69.61& 63.48 \\ 
        \bottomrule
    \end{tabular}\vspace{-3mm}
    }
    \caption{{\small{Impact of scaling up teacher size on model performance trained on 7B tokens. Except for teacher size, we use the same training configurations for the same size of student models. }}}
    \vspace{-2mm}
    \label{ablation:teacher}
\end{table*}

\vspace{-6pt}
\subsection{{Extension to Qwen models}}
\vspace{-6pt}
{We further demonstrate that the proposed initialization and distillation strategy is not limited to Llama models but also applicable to other popular model families such as Qwen \cite{hui2024qwen2}, which we refer to as Zebra-Qwen models. We choose Qwen-2.5-0.5B-Instruct and Qwen-2.5-1.5B-Instruct as the base models. For the 0.5B model, we use the 1.5B mode as the teacher and replace 4 layers to MLA and 20 layers to Mamba2, resulting in 6.25\% KV cache size. For the 1.5B model, we use itself as the teacher model and replace 14 layers to MLA and 14 layers to Mamba2, with a KV size of 12.5\% of the original target model. The performance is illustrated in Table \ref{tab:qwen_results}. The conclusion remains the same as Llama models - The 0.5B Zebra-Qwen model achieves 3.65\% higher accuracy than the base model and 1.5B Zebra-Qwen model only experiences 0.7\% performance degradation with 8$\times$ KV cache compression.}

\begin{table*}[t]
     \setlength\extrarowheight{2pt}
    \centering
    \scalebox{0.63}{
    \begin{tabular}{l|ccc|ccccccccccc}
        \toprule
        Model and Setting & Teacher & Tokens & KV Size & ARC & ARE & HS & MM & OB & PI & RA & WG & Avg. \\        \midrule
        \midrule
        Qwen2.5-0.5B-Inst & - & - & 100\% &33.11	&59.05	&52.26	&45.86	&34.2	&70.62	&32.06	&56.27 & \textbf{47.93} \\
        \hdashline
        \rowcolor{gray!20}
        Zebra-Qwen-0.5B, 4MLA-20M2 & 1.5B & 7B & 6.25\% & 38.74	&66.92	&50.83	&38.43	&37.2	&69.91	&32.34	&55.09 &48.68\\ 
        \bottomrule
        \toprule
         Qwen2.5-1.5B-Inst & - & - & 100\% &47.01 &75.84	&68.24	&60.13	&41	&76.01	&37.8	&62.67 & \textbf{58.59} \\
        \hdashline
        \rowcolor{gray!20}
        Zebra-Qwen-1.5B, 14MLA-14M2 & 1.5B & 7B & 6.25\% & 48.63 &75.17	&67.64	&53.87	&41.6	&75.73	&38.66	&64.01 & 58.16 \\ 
        \bottomrule
    \end{tabular}
    }
    \vspace{-3pt}
    \caption{{\small{Zero-shot evaluation on the LM Harness Eval benchmark for Zebra-Qwen models}.}}
    \label{tab:qwen_results}
    \vspace{-15pt}
\end{table*}

\vspace{-6pt}

\subsection{{Extended Long-Context Evaluation}}

\label{appendix:longcontext}

\begin{wraptable}{R}{0.48\textwidth}
    \centering
    \vspace{-0.4in}
    \scalebox{0.8}{
    \begin{tabular}{l|c|ccc}
        \toprule
        \textbf{Model} & \textbf{KV\%} & \textbf{4K} & \textbf{8K} & \textbf{16K} \\ 
        \midrule \midrule
        MambaInLlama-1B & 50\% & 38.75 & 21.55 & 3.88 \\
        Zebra-Llama-1B & 3.91\% & 35.75 & \textbf{24.80} & \textbf{13.37} \\ 
        \hdashline
        MambaInLlama-3B & 50\% & 41.71 & 22.62 & 0.88 \\
        Zebra-Llama-3B & 4.69\% & \textbf{58.69} & \textbf{38.24} & \textbf{9.97} \\ 
        \bottomrule
    \end{tabular}
    }
    \caption{\small{RULER benchmark results at 4K, 8K, and 16K context lengths.}}
    \label{tab:longcontext_ruler}
    \vspace{-3mm}
\end{wraptable}
 {
For the long-context evaluation of our models, a few factors should be considered.
First, assessing long-context performance requires models that have been trained on datasets with extended sequence lengths.
However, due to efficiency and resource constraints, our current models were trained with a maximum sequence length of {2048 tokens}.
Such a training configuration is generally not optimal for long-context benchmarks like \textbf{RULER}~\cite{hsieh2024ruler}.
Despite this limitation, we conducted RULER benchmark evaluations at \textbf{4K}, \textbf{8K}, and \textbf{16K} context lengths and the results are summarized in Table~\ref{tab:longcontext_ruler}. We observe that, at the 3B scale, \textbf{Zebra-Llama surpasses MambaInLlama across all context lengths}, while at the 1B scale it performs on par at 4K and exceeds MambaInLlama at 8K and 16K. 
Importantly, these gains are achieved with a substantially smaller memory footprint—\textbf{Zebra-Llama uses only $\sim$5\% of the KV cache}, whereas MambaInLlama requires nearly ten times more KV memory. 

As future work, we aim to extend our models to support longer context lengths and reasoning. 
}

\vspace{-8pt}
\section{Conclusion}
    \vspace{-10pt}
In this work, we addressed the growing need for efficient LLMs by proposing a practical and scalable framework for composing hybrid models from existing pre-trained Transformers. Motivated by the cost and environmental impact of retraining large models for downstream use, we introduced \ours, a family of 1B, 3B, and 8B hybrid models built using SSMs and MLA layers. We developed an effective initialization scheme and a post-training knowledge transfer pipeline that enabled these models to inherit capabilities from larger teacher models with minimal additional training.
Our approach significantly reduced memory while preserving the accuracy of strong baselines. 
\vspace{-12pt}
\paragraph{Limitations and Future Work} 
Our work establishes several promising directions for future research. A primary next step is to move beyond a single family of base models and explore hybridization strategies across diverse LLM architectures, particularly incorporating modular frameworks like Mixture-of-Experts (MoE). Scaling our training pipeline to larger models and extending the approach to reasoning-intensive architectures remain critical frontiers. Furthermore, addressing the reliance on strong teacher models is essential. In constrained scenarios, investigating alternative knowledge transfer strategies—such as teacher-free or efficient self-distillation—will be key to understanding how hybrid architectures can effectively learn with reduced supervision.


\printbibliography








\newpage
\appendix

\section{More Experimental Details for \ours}

\subsection{Structured MLA Initialization} \label{app:mla_init}

\begin{algorithm}[h]
\begin{lstlisting}
# MHA weights: W_Q, W_K, W_V
# MLA weights: W_DQ, W_UQ, W_QR, W_DKV, W_UK, W_KR, W_UV

# Initialization of W_DQ, W_UQ, and W_QR
U_q, sigma_q, V_q = svd(W_Q)
W_DQ = U_q
W_UQR_bar = (sigma_q @ V_q).view(r_q, n_h, d_h)
W_UQ = W_UQR_bar[:, :, :d_qk].view(r_q, n_h*d_qk)
W_QR = W_UQR_bar[:, :, -d_r:].view(r_q, n_h*d_r)

# Initialization of W_DKV, W_UK, W_KR, W_UV
U_kv, sigma_kv, V_kv = svd(torch.cat((W_K, W_V), -1))
W_DKV = U_kv
W_K_avg = W_K.view(d, n_kv, d_h).mean(1)
W_KR = W_K_avg[:, -d_r:]

W_UKV = sigma_kv @ V_kv
W_UK_bar = W_UKV[:, :d_h*n_kv].view(r_kv, n_kv, d_h)
W_UK = W_UK_bar[:,:,:d_qk].view(r_kv, n_kv*d_qk)
W_UV = W_UKV[:, d_h*n_h:]
\end{lstlisting}
\caption{Python-like pseudocode of the proposed SVD initialization for MLA.}
\label{alg:pseudocode}
\end{algorithm}

Our SVD-based MLA layer initialization follows the methodology outlined in X-EcoMLA \cite{li2025x} for Multi-Head Attention (MHA). However, for Generalized Question Answering (GQA) models like the Llama 3 series, our approach diverges slightly by keeping the number of key/value heads from the base model for MLA while X-EcoMLA forces the number of key/value heads to be the same as the number of query head. With such modification, we notice trivial performance difference but with slightly fewer number of parameters. 

Essentially, the matrices from the base MHA/GQA module could be expressed as $W^Q\in\mathbb{R}^{d\times n_hd_h}$, and $W^K, W^V\in\mathbb{R}^{d\times n_{kv}d_h}$, where $d$ denotes the hidden dimension, $n_h$ denotes the number of attention heads, $n_{kv}$ denotes the number of key/value heads. For MHA, we have $n_h=n_{kv}$ while for GQA we have $n_h > n_{kv}$. 

All the matrices we need to initialize in MLA could be expressed as $W^{DKV}\in\mathbb{R}^{d\times r_{kv}}$, $W^{UK}\in\mathbb{R}^{r_{kv}\times n_{kv}d_{qk}}$, $W^{UV}\in\mathbb{R}^{r_{kv}\times n_{kv}d_{v}}$, $W^{DQ}\in\mathbb{R}^{d\times r_q}$, $W^{UQ}\in\mathbb{R}^{r_q\times n_hd_{qk}}$, $W^{KR}\in\mathbb{R}^{d\times d_{r}}$, and $W^{QR}\in\mathbb{R}^{r_q\times n_hd_{r}}$, where $r_{kv}$ represents the latent dimension for key/value, $d_{qk}$ denotes the head dimension for query/key, $d_v$ denotes the head dimension for value, $r_q$ denotes the latent dimension for query, and $d_r$ represents the dimension for RoPE embeddings. For all the experiments, we keep $d_v=d_h$ and $d_{qk}+d_{r}=d_h$. 

\paragraph{Initialization of $W^{DQ}$, $W^{UQ}$, and $W^{QR}$} Given that query compression in MLA can be viewed as a low-rank approximation of attention layers, we initially perform SVD on the pre-trained weight matrix $W^Q$:
\begin{equation}
     W^Q = U_q \Sigma_q V^T_q,
\end{equation}
where $U_q \in \mathbb{R}^{d \times r_q}$, $\Sigma_q \in \mathbb{R}^{r_q \times r_q}$, and $V_q \in \mathbb{R}^{d_h n_h \times r_q}$. For constructing the up-projection matrices, we reshape the product $\Sigma_q V^T_q$ to form $\overline{W}^{UQR} \in \mathbb{R}^{r_q \times n_h \times d_h}$ and subsequently split this tensor at the last dimension into $W^{UQ}$, containing the first $d_{qk}$ elements, and $W^{QR}$, containing the remaining $d_r$ elements. The down-projection matrix $W^{DQ}$ is directly assigned with $U_q$. This method of initial assignment is formulated as:
\begin{equation}
    W^{DQ} = U_q, \;\;W^{UQ} = \text{reshape}(\overline{W}^{UQR}[:,:,:d_{qk}]), \;\;
    W^{QR} = \text{reshape}(\overline{W}^{UQR}[:,:,-d_r:]),
\end{equation}
where the function $\text{reshape}(.)$ is used to integrate the last two dimensions of the specified tensor.

\paragraph{Initialization of $W^{DKV}$, $W^{UK}$, $W^{UV}$, and $W^{KR}$} The initialization of the MLA weights associated with keys and values is more complicated because of the decoupled RoPE mechanism. First, a joint SVD is performed on the concatenated $W^K$ and $W^V$:
\begin{equation}
    [W^K, W^V] = U_{kv} \Sigma_{kv} V^T_{kv}, 
\end{equation}
where $U_{kv} \in \mathbb{R}^{d \times r_{kv}}$, $\Sigma_{kv} \in \mathbb{R}^{r_{kv} \times r_{kv}}$, and $V_{kv} \in \mathbb{R}^{2 d_h n_{kv} \times r_{kv}}$. For the down-projection matrix $W^{DKV}$, we directly set it to $U_{kv}$. To derive the up-projection matrix $W^{UV}$ and $W^{UK}$, we first set $W^{UKV} = \Sigma_{kv} V^T_{kv}$. Since we have $d_v=d_h$, we simply extract the last $d_h n_{kv}$ columns of $W^{UKV}$ as $W^{UK}$. For $W^{UK}$, we first extract the first $d_h n_{kv}$ columns of $W^{UKV}$ and reshape them into $\overline{W}^{UK} \in \mathbb{R}^{r_{kv} \times n_{kv} \times d_h}$. Then, we select the first $d_{qk}$ elements along the last dimension of $\overline{W}^{UK}$ and reshape it back to obtain $W^{UK}$. In general, this process can be expressed as:
\begin{equation}
    W^{DKV} = U_{kv}, \;\; W^{UV} = W^{UKV}[:,:n_{kv}d_{h}], \;\;
    W^{UK} = \text{reshape}(\overline{W}^{UK}[:,:,:d_{qk}]).
\end{equation}  
In the final step, the initialization of the RoPE key embedding matrix $W^{KR}$ requires a distinct approach, given that all attention heads in MLA utilize the identical RoPE key embedding. First, the average key projection matrix $W^K_{avg} \in \mathbb{R}^{d \times d_h}$ is calculated for all attention heads. Then, the final $d_r$ columns are extracted for initializing $W^{KR}$, formulated as follows: 
\begin{equation}
    W^{KR}=W^K_{avg}[:, -d_{r}:].
\end{equation}
The detailed initialization algorithm can be found in Algorithm \ref{alg:pseudocode}.

\subsection{Structured Mamba2 Initialization} \label{app:mamba_init}
The structured initialization process of Mamba2 layers follows precisely to the solution of MambaInLlama \cite{wang2025mamballamadistillingaccelerating}. As outlined in Section \ref{Initialization}, excluding the softmax operation in attention shows a direct one-to-one mapping between $B_t$, $x_t$, and $C_t$ in linear RNN and $K_t$, $V_t$, and $Q_t$ in attention layers. In the Mamba2 framework, $B_t$, $x_t$, and $C_t$ for the continuous-time SSM are derived from input $o_t$ via passing through a MLP followed by a 1d convolution layer. The MLP is replaced directly with pre-trained transformer layer weights as follows:
\begin{equation}
    x_t=W^Vo_t\in \mathbb{R}^{b\times n_{kv}h_d}, B_t = W^Ko_t \in \mathbb{R}^{b\times n_{kv}h_d}, C_t = W^Qo_t\in \mathbb{R}^{b\times n_{h}h_d},
\end{equation}
where $b$ signifies the batch size. Subsequent to this, $x$, $B$, and $C$ are processed through the 1d convolutional layer for temporal fusion before undergoing discretization in Mamba SSM. It is important to highlight that for GQA and MQA scenarios where $n_{kv} < n_q$, $x_t$ and $B_t$ are replicated after the convolution to ensure that $x_t$, $B_t$, and $C_t$ share identical dimensions. Other parameters, such as $A$ and $\Delta_t$, adhere to the original initialization procedure within the Mamba2 layers.

\subsection{Model Structure Details} \label{app:structure}

Detailed architectures of our 1B, 3B, and 8B models, as presented in Table \ref{tab:main_results}, are comprehensively outlined in Table \ref{appendix:model_structure}. This includes MLA layer selections, parameters specific to the MLA layers, and the overall model size. Layers not categorized as MLA are Mamba2 layers, which follows the configuration used in MambaInLlama \cite{wang2025mamballamadistillingaccelerating}.

\begin{table*}[t]
    \setlength\extrarowheight{2pt}
    \centering
    \scalebox{0.78}{
    \begin{tabular}{c|cccccc}
        \toprule
        Method & MLA Indices & $r_{kv}$ & $r_{q}$ & $d_{qk}$ & $d_r$ & Model Size \\    \midrule \midrule
        \ours-1B, 8MLA8M2 &{[}0,2,4,8,10,12,14{]} & 128 & 1344 & 32 & 32 & 1.27B\\
        \ours-1B, 6MLA10M2 &{[}0,2,5,8,11,14{]} & 128 & 1344 & 32 & 32 & 1.28B \\ 
        \ours-1B, 4MLA12M2 &{[}0,5,10,14{]} & 128 & 1344 & 32 & 32 & 1.28B \\
        \hdashline
        \ours-3B, 14MLA14M2 &{[}0,2,4,6,8,10,12,14,16,18,20,22,24,{26}{]}& 128 &  1536 & 64 & 64 & 3.27B \\
        \ours-3B, 8MLA20M2 &{[}0,4,8,12,16,20,24,{26}{]} & 128 &  1536 & 64 & 64 & 3.36B \\
        \ours-3B, 6MLA22M2 &{[}{0,5,10,16,21,26}{]} & 128 &  1536 & 64 & 64 & 3.39B \\ \hdashline
        \ours-8B, 16MLA16M2 &{[}0,2,4,6,8,10,12,14,16,18,20,22,24,26,28,{30}{]}& 160 &  2048 & 64 & 64 & 8.19B\\
        \ours-8B, 8MLA24M2 &{[}{0,4,8,12,16,20,25,30}{]} & 160 &  2048 & 64 & 64 & 8.38B\\
        \bottomrule
    \end{tabular}
    }
    \caption{\small{Configurations of \ours models' architecture.}}
    \label{appendix:model_structure}
\end{table*}

\subsection{Training Details} \label{app:hyper_param}
In Table \ref{appendix:training_details}, we present the training configurations for our \ours series models, including the number of tokens, batch size, learning rate, and total training time. All experiments are conducted on a single node equipped with eight AMD MI300 GPUs, each featuring 192GB of memory. We apply a learning rate warmup over the first $1\%$ of training data, followed by cosine annealing. The models are optimized using AdamW, with hyperparameters set to $\beta = (0.9, 0.8)$. Additionally, all models process input sequences of length 2048 through sample packing.

\begin{table*}[t]
    \setlength\extrarowheight{2pt}
    \centering
    \scalebox{1}{
    \begin{tabular}{c|cccccc}
        \toprule
        Stage & Model & Teacher & Tokens & BS & LR & Time (H) \\    \midrule \midrule
        ILD & \ours-1B-MLA & Llama3.2-1B & 1.36B & 96 & 8e-5 & 1.8\\
        ILD & \ours-1B-Mamba2 & Llama3.2-1B & 1.36B & 96 & 8e-5 & 1.7\\
        ILD & \ours-3B-MLA & Llama3.2-3B & 1.36B & 96 & 8e-5 & 3.9 \\
        ILD & \ours-3B-Mamba2 & Llama3.2-3B & 1.36B & 96 & 8e-5 & 4.4 \\
        ILD & \ours-8B-MLA & Llama3.1-8B & 1.36B & 48 & 4e-5 & 9.2\\
        ILD & \ours-8B-Mamba2 & Llama3.1-8B & 1.36B & 48 & 4e-5 & 10.3\\
        \hdashline
        SFT & \ours-1B & Llama3.1-8B & 5.44B & 192 & 8e-5 & $\approx 13.7$\\
        SFT & \ours-3B & Llama3.1-8B & 7.44B & 96 & 4e-5 & $\approx 31.2$\\
        SFT & \ours-8B & Llama3.1-8B & 9.44B & 64 & 4e-5 & $\approx 78.1$\\
        \hdashline
        DPO & \ours-1B & N/A & 0.2B & 32 & 5e-7 & $\approx 0.5$\\
        DPO & \ours-3B & N/A & 0.2B & 32 & 5e-7 & $\approx 1.2$\\
        DPO & \ours-8B & N/A & 0.2B & 32 & 5e-7 & $\approx 2.3$\\
        \bottomrule
    \end{tabular}
    }
    \caption{\small{Training hyper-parameters of \ours models.}}
    \label{appendix:training_details}
\end{table*}

\subsection{Evaluation Details}\label{app:eval_details}
We evaluate all models using the \href{https://github.com/EleutherAI/lm-evaluation-harness}{lm-evaluation-harness library} (commit from the big-refactor branch) following the task-specific few-shot configurations defined by the Open LLM Leaderboard. For zero-shot evaluation, we report performance across a broad suite of language understanding tasks: MMLU, HellaSwag, PIQA, ARC-Easy, ARC-Challenge, Winogrande, OpenBookQA, and RACE. Evaluations are performed using the command-line interface with ROCm-enabled devices and a batch size of 16. For few-shot runs targeting leaderboard comparisons, we use the officially recommended number of shots per task (e.g., 25-shot for ARC-Challenge, 10-shot for HellaSwag, 5-shot for Winogrande and MMLU, 0-shot mc2 for Truthful-QA(TQ)) ~\cite{lin2021truthfulqa}. We report average scores across tasks, following the same protocol as prior work. 

\section{\smart Layer Selection Algorithms}
\label{apndx:SMART}



\begin{algorithm}[t]
\caption{Pseudo code: SMART---Structured MLA Layer Selection via Sensitivity Scores}\label{alg:layer_select}
\label{alg:smart}
\begin{algorithmic}[1]
\Require Sensitivity scores $\{s_1, s_2, \ldots, s_L\}$, number of MLA layers $N$
\Ensure Selected MLA layer indices $\{L_1^{\text{MLA}}, \ldots, L_N^{\text{MLA}}\}$

\State \textbf{// Terminal Preservation}
\State Divide $L$ layers into $N$ equal partitions
\State $L_1^{\text{MLA}} \gets$ highest-sensitivity layer in first partition of layers $\left\{i, i\in \left[1, \left\lfloor \frac{L}{N} \right\rfloor\right]\right\}$
\State $L_N^{\text{MLA}} \gets$ highest-sensitivity layer in last partition of layers $\left\{i, i\in \left[L-\left\lfloor \frac{L}{N}  \right\rfloor+1, L\right]\right\}$

\State \textbf{// Near-Uniform Intermediate Distribution}
\State \textit{// Define Valid Intermediate Layer Range}
\State Let $r_{\text{start}} \gets L_1^{\text{MLA}} + 1$
\State Let $r_{\text{end}} \gets L_N^{\text{MLA}} - 1$
\State Let $R \gets$ list of candidate intermediate layers in $[r_{\text{start}}, r_{\text{end}}]$

\State \textit{// Compute Allowable Gap Bounds}
\State Let $T \gets L_N^{\text{MLA}} - L_1^{\text{MLA}} - N + 1$
\State Let $g_{\min} \gets \left\lfloor \frac{T}{N - 1} \right\rfloor$
\State Let $g_{\max} \gets \left\lceil \frac{T}{N - 1} \right\rceil$

\State \textit{// Enumerate All Valid Configurations}
\State Initialize empty list of configurations $\mathcal{C} \gets []$
\ForAll{combinations of $N - 2$ layers from $R$}
    \State Sort the selected layers in ascending order to form $C_j$
    \State Let $G_j \gets$ list of gaps between consecutive layers in $\{L_1^{\text{MLA}}\} \cup C_j \cup \{L_N^{\text{MLA}}\}$
    \If{all gaps in $G_j$ satisfy $g_{\min} \leq \text{gap} \leq g_{\max}$}
        \State Append $C_j$ to $\mathcal{C}$
    \EndIf
\EndFor
\State \textbf{// Maximal Sensitivity Scores}
\State $C^* \gets \arg\max_{C_j} \sum_{i \in C_j} s_i$
\State \Return $\{L_1^{\text{MLA}}\} \cup C^* \cup \{L_N^{\text{MLA}}\}$

\end{algorithmic}
\end{algorithm}

We provide the pseudo code for SMART in Algorithm~\ref{alg:layer_select}. Besides, we provide three examples for the layer selection process of our \ours 1B models following \smart. The examples are based on the sensitivity analysis shown in Figure \ref{fig:sensitivity}, whose actually values are listed in Table \ref{app:scores}.

\begin{table}[t]
    \centering
    
    \begin{tabular}{|c|c|c|c|}
        \hline
        Layer Index & Sensitivity Score & Layer Index & Sensitivity Score \\
        \hline
        0 & 1185.06 & 8 & 238.1\\
        1 & 382.73 & 9 & 120.56\\
        2 & 480.68 & 10 & 323.23\\
        3 & 350.95 & 11 & 228.9\\
        4 & 196.03 & 12 & 168.69\\
        5 & 367.82 & 13 & 233.87\\
        6 & 250.45 & 14 & 624.03\\
        7 & 114.44 & 15 & 361.47\\
        \hline
    \end{tabular}

    \vspace{0.1in}
    \caption{\small{{Concrete} sensitivity score in Figure \ref{fig:sensitivity}.}}
    \label{app:scores}
\end{table}

\begin{tcolorbox}[colframe=black,colback=white,boxrule=1pt]
\begin{minipage}{\textwidth}
\paragraph{Example 1: \ours-1B with $N=4$} 

\begin{itemize}
    \item Terminal Preservation: $L^{MLA}_1=0$, $L^{MLA}_N=14$
    \item Near-Uniform Intermediate Distribution: 
    \begin{itemize}
        \item Define intermediate layer range: $r_{start}=1$, $r_{end}=13$, $R=\{1,2,\ldots,13\}$.
        \item Compute allowable gap bounds: $T=11$, $g_{min}=3$, $g_{max}=4$.
        \item Enumerate valid Configurations: $\mathcal{C}=\{\{4,9\}, \{5,9\}, \{5,10\}\}$.
    \end{itemize}
    \item Maximal Sensitivity Scores: 
    \begin{itemize}
        \item Calculate total sensitivity score: 
        \begin{itemize}
            \item $s_4+s_9=316.59$,
            \item $s_5+s_9=488.38$,
            \item $s_5+s_{10}=691.05$
        \end{itemize}
        \item Layers with maximal score: $C^*=\{5,10\}$.
        \item Return $\{0,5,10,14\}$.
    \end{itemize}
\end{itemize}
\end{minipage}
\end{tcolorbox}

\begin{tcolorbox}[colframe=black,colback=white,boxrule=1pt]
\begin{minipage}{\textwidth}
\paragraph{Example 2: \ours-1B with $N=6$} 

\begin{itemize}
    \item Terminal Preservation: $L^{MLA}_1=0$, $L^{MLA}_N=14$
    \item Near-Uniform Intermediate Distribution: 
    \begin{itemize}
        \item Define intermediate layer range: $r_{start}=1$, $r_{end}=13$, $R=\{1,2,\ldots,13\}$.
        \item Compute allowable gap bounds: $T=9$, $g_{min}=1$, $g_{max}=2$.
        \item Enumerate Valid Configurations: $\mathcal{C}=\{$ $\{2,5,8,11\}$, $\{3,5,8,11\}$, $\{3,6,8,11\}$, $\{3,6,9,11\}$, $\{3,6,9,12\}\}$.
    \end{itemize}
    \item Maximal Sensitivity Scores: 
    \begin{itemize}
        \item Calculate total sensitivity score: 
        \begin{itemize}
            \item $s_2+s_5+s_8+s_{11}=1315.5$, 
            \item $s_3+s_5+s_8+s_{11}=1185.8$, 
            \item $s_3+s_6+s_8+s_{11}=1068.4$,
            \item $s_3+s_6+s_9+s_{11}=950.86$,
            \item $s_3+s_6+s_9+s_{12}=890.65$. 
        \end{itemize}
        
        \item Layers with maximal score: $C^*=\{2,5,8,11\}$.
        \item Return $\{0,2,5,8,11,14\}$.
    \end{itemize}
\end{itemize}
\end{minipage}
\end{tcolorbox}

\begin{tcolorbox}[colframe=black,colback=white,boxrule=1pt]
\begin{minipage}{\textwidth}
\paragraph{Example 3: \ours-1B with $N=8$} 

\begin{itemize}
    \item Terminal Preservation: $L^{MLA}_1=0$, $L^{MLA}_N=14$
    \item Near-Uniform Intermediate Distribution: 
    \begin{itemize}
        \item Define intermediate layer range: $r_{start}=1$, $r_{end}=13$, $R=\{1,2,\ldots,13\}$.
        \item Compute allowable gap bounds: $T=7$, $g_{min}=1$, $g_{max}=1$.
        \item Enumerate Valid Configurations: $\mathcal{C}=\{\{2,4,6,8,10,12\}\}$.
    \end{itemize}
    \item Only one valid configuration. Return $\{0,2,4,6,8,10,12,14\}$.
\end{itemize}
\end{minipage}
\end{tcolorbox}

\section{Comparison with Pre-training Methods}
\label{sec:main_results_pretrain}
In Table~\ref{tab:main_results_pretrain}, we include more comparisons of our \ours with other hybrid models based on pre-training instead of distillation. It can be observed that our method could achieve competitive performance as the pre-trained baselines $214\times-500\times$ fewer training tokens, demonstrating our advantage in the training efficiency.
For instance, \ours (16MLA–16M2) matches or exceeds the accuracy of Mamba-2-Hybrid (8.66B, 3.5T tokens) and SAMBA (1.7B, 3.5T tokens), while using only 11B tokens and smaller model sizes. This highlights the efficiency and practicality of our post-training hybrid composition strategy, as it achieves high performance with a fraction of the training budget required by scratch-trained models.

\begin{table*}[t]
     \setlength\extrarowheight{2pt}
    \centering
    \scalebox{0.65}{
    \begin{tabular}{l|cc|cccccccccc}
        \toprule
        Model and Setting  & Tokens & Size  & ARC & ARE & HS & MM & OB & PI & RA & WG  \\  
        \midrule
        \midrule
        Mamba-2-Hybrid  & 3.5T & 8.66B  & -~-~-~-/47.7 &77.23/-~-~-~-&	-~-~-~-/77.68&	51.46&	-~-~-~-/42.8&	79.65/-~-~-~-&	39.71&	71.27\\ 
        SAMBA  & 3.5T & 1.7B  & 48.21/-~-~-~- & 79.25/-~-~-~-&	49.74/-~-~-~-&	48.01&	37.20/-~-~-~-&	77.10/-~-~-~- &	-&	72.93 \\ 
        SAMBA  & 3.5T & 1.3B  & - &58.21/-~-~-~-&	-~-~-~-/54.73&	-&	-&	72.36/-~-~-~-&	-&	55.72 \\         
        Hymba  & 1.5T & 1.5B  & 42.32/-~-~-~- &74.54/-~-~-~-&	53.55/-~-~-~-&	-&	-&	76.66/-~-~-~-&	-&	66.61 \\         
        \hdashline
        \rowcolor{gray!20}
        \ours, 8MLA-8M2  & 7B & 1.27B  & 39.68/42.83&	72.35/67.3&	45.26/60.59&	38.99&	30.2/41.6&	72.91/73.29&	38.56&	61.33 \\ 
        \rowcolor{gray!20}
        \ours, 14MLA-14M2  & 9B & 3.27B  & 50/52.65&	80.47/76.35&	53.52/72.43&	51.97&	31.8/44.4&	76.61/76.99&	46.99&	67.8 \\ 
        \rowcolor{gray!20}
        \ours, 16MLA-16M2  & 11B & 8.19B  & 57.17/58.96&	83.59/79.92&	57.82/77.73&	57.18&	35.20/44.6&	79.65/80.2&	48.71&	72.38\\        
        \bottomrule

    \end{tabular}
    }
    \caption{Comparing our \ours with state-of-the-art hybrid models that are trained from scratch. We report two accuracy scores for each model, i.e, accuracy and normalized accuracy (acc / acc\_norm). The missing results (`-') in the table means the results are not reported in original paper. }
    \label{tab:main_results_pretrain}
\end{table*}

\section{More Inference Evaluations}

\begin{figure}[t]
\centering
  \includegraphics[width=0.95\columnwidth]{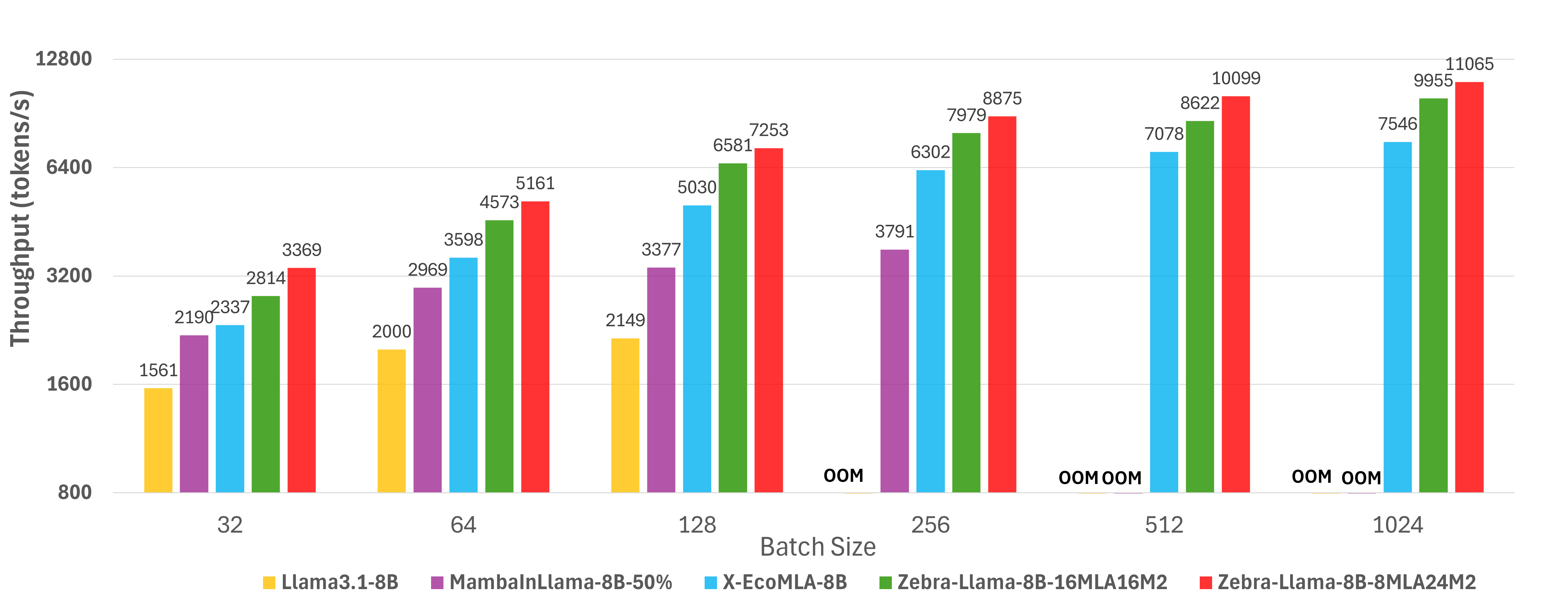}
  \vspace{-5pt}
  \caption{\small{Inference throughput vs. batch size of various 8B-size models. We measure the throughput with output sequence length 8192.} }
  \label{fig:infer_8B_batch_throughput}
  \vspace{-5pt}
\end{figure}
\begin{figure}[t]
\centering
  \includegraphics[width=0.95\columnwidth]{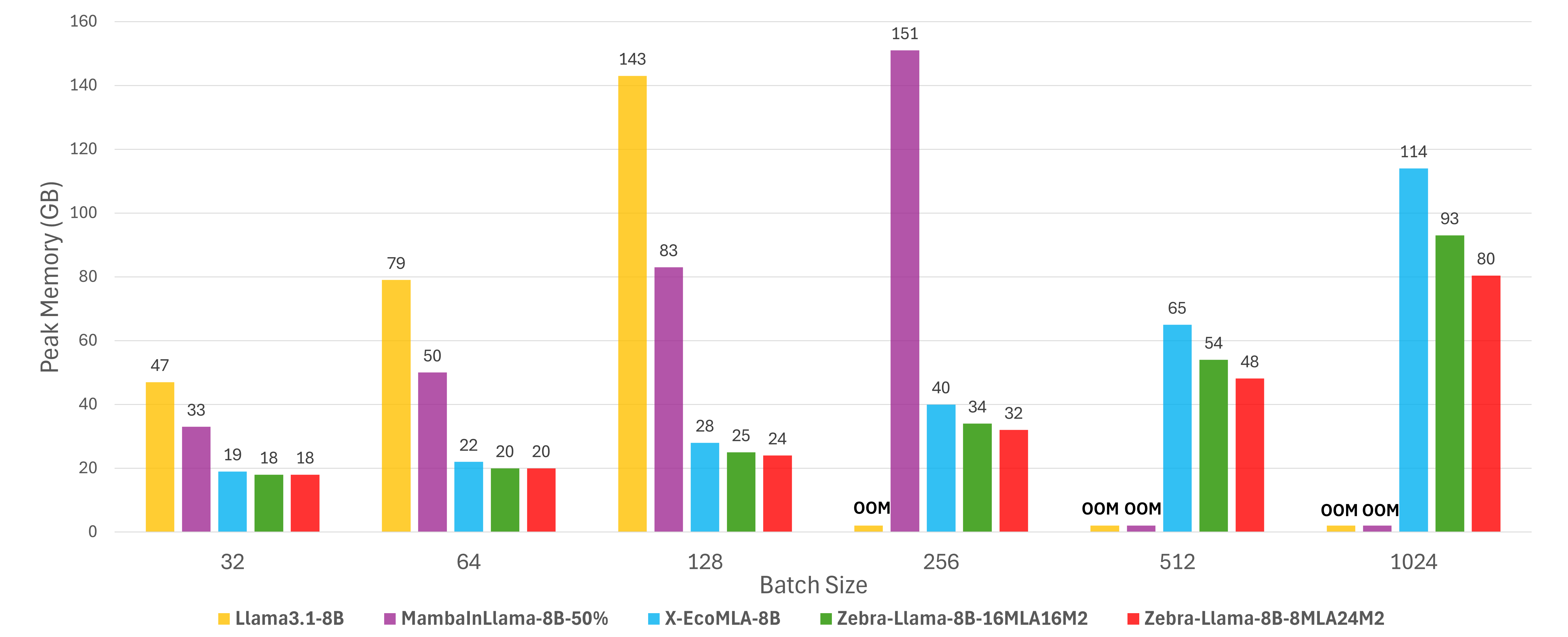}
  \vspace{-5pt}
  \caption{\small{Peak memory usage during inference  vs. batch size of various 8B-size models. We measure the memory footprint with output sequence length 8192.} } \label{fig:infer_8B_batch_memory}
  \vspace{-5pt}
\end{figure}

{In Figure \ref{fig:infer_8B_batch_throughput} and \ref{fig:infer_8B_batch_memory}, we include more inference evaluations for 8B models iunder a setting where the prompt length is fixed at 1, the output length at 8192, and the batch size gradually increases from 32 to 1024. All measurements are recorded on a single AMD MI300 GPU with 192GB of memory with Hipgraph. 
Our \ours models demonstrate significant throughput gains compared to Llama, X-EcoMLA (pure MLA), and MambaInLlama (hybrid GQA-Mamba). Additionally, the KV cache compression in our \ours models translates to notable memory savings, as illustrated in Figure \ref{fig:infer_8B_batch_memory}. Models incorporating GQA layers experience a sharp increase in memory usage as the batch size grows. In contrast, MLA-based approaches, such as X-EcoMLA and our \ours models, demonstrate superior memory efficiency.}

\section{ {Extended Few-Shot Evaluation Results}}
\label{appendix:fewshot}

 {In the main text, we focused on reporting the few-shot performance of the \textbf{8B model} due to space constraints. 
However, for completeness, we include the full few-shot evaluation results for the \textbf{1B} and \textbf{3B} model variants as well.  
As shown below, our \textbf{Zebra-Llama} models maintain competitive few-shot performance across multiple benchmarks even at smaller scales, demonstrating that our hybrid \textit{MLA–Mamba} design preserves strong generalization while significantly reducing KV cache requirements.}
 {At both 1B and 3B scales, the proposed \textit{Zebra-Llama} configurations maintain or even exceed the baseline performance on several tasks, while operating with drastically reduced KV cache memory (as low as 2.9\%). 
This demonstrates that our hybrid \textbf{MLA–Mamba2} architecture effectively balances efficiency and few-shot reasoning ability even at smaller model sizes.}

\begin{table}[h!]
\centering
\resizebox{\textwidth}{!}{
\begin{tabular}{lccccccc}
\toprule
\textbf{Model and Setting} & \textbf{KV\%} & \textbf{Avg.} & \textbf{ARC(25)} & \textbf{HS(10)} & \textbf{MMLU(5)} & \textbf{WG(5)} & \textbf{TQ(0)} \\
\midrule
Llama3.2-1B-Instruct & 100\% & 49.98 & 41.38 & 59.80 & 45.48 & 59.35 & 43.88 \\
MambaInLlama-1B-50\%* & 50\% & 48.60 & 42.32 & 60.46 & 35.55 & 59.35 & 45.31 \\
X-EcoMLA-1B & 9.37\% & 47.97 & 41.04 & 56.13 & 35.32 & 60.77 & 46.59 \\
Llamba-1B & 0\% & 47.57 & 41.72 & 60.34 & 31.88 & 60.69 & 43.20 \\
\textbf{Zebra-Llama-1B, 8MLA-8M2 (Ours)} & \textbf{7.80\%} & \textbf{49.68} & \textbf{45.56} & \textbf{59.44} & \textbf{37.81} & \textbf{60.77} & \textbf{44.80} \\
\textbf{Zebra-Llama-1B, 6MLA-10M2 (Ours)} & \textbf{5.86\%} & \textbf{49.06} & \textbf{44.03} & \textbf{59.22} & \textbf{36.06} & \textbf{60.54} & \textbf{45.46} \\
\bottomrule
\end{tabular}
}
\caption{\small{\textbf{Few-Shot Performance for 1B Models.} 
Each model is evaluated on five standard few-shot benchmarks: ARC (25-shot), HellaSwag (10-shot), MMLU (5-shot), Winogrande (5-shot), and TruthfulQA (0-shot). 
KV\% denotes the relative KV-cache size compared to the dense Llama baseline.}}
\label{tab:fewshot_1b}
\end{table}

\begin{table}[h!]
\centering
\label{tab:fewshot_3b}
\resizebox{\textwidth}{!}{
\begin{tabular}{lccccccc}
\toprule
\textbf{Model and Setting} & \textbf{KV\%} & \textbf{Avg.} & \textbf{ARC(25)} & \textbf{HS(10)} & \textbf{MMLU(5)} & \textbf{WG(5)} & \textbf{TQ(0)} \\
\midrule
Llama3.2-3B-Instruct & 100\% & 60.54 & 52.39 & 73.51 & 59.71 & 67.32 & 49.75 \\
MambaInLlama-3B-50\% & 50\% & 61.18 & 51.88 & 74.58 & 52.31 & 67.64 & 59.51 \\
X-EcoMLA-3B* & 9.37\% & 57.50 & 49.49 & 69.20 & 52.26 & 66.69 & 49.86 \\
Llamba-3B & 0\% & 58.01 & 50.09 & 74.21 & 49.87 & 70.09 & 45.79 \\
\textbf{Zebra-Llama-3B, 14MLA-14M2 (Ours)} & \textbf{4.69\%} & \textbf{60.12} & \textbf{53.67} & \textbf{71.30} & \textbf{51.05} & \textbf{67.64} & \textbf{56.94} \\
\textbf{Zebra-Llama-3B, 8MLA-20M2 (Ours)} & \textbf{2.86\%} & \textbf{58.49} & \textbf{54.52} & \textbf{70.44} & \textbf{46.43} & \textbf{65.98} & \textbf{55.07} \\
\bottomrule
\end{tabular}
}
\caption{\small{\textbf{Few-Shot Performance for 3B Models.} 
Same benchmarks and settings as Table~\ref{tab:fewshot_1b}.}}

\end{table}

\subsection{Extending Evaluation to More Challenging Benchmarks}
\label{appendix:challenging_benchmarks}
 {
In this section, we evaluate our models on more challenging benchmarks for assessing deeper reasoning capabilities, particularly given the architectural innovations in our hybrid design. 
Here, we report our results on {GSM8K} for mathematical reasoning, {GPQA}~\cite{rein2024gpqa} for graduate-level scientific question answering and RACE~\cite{lai2017race}, a well-established benchmark for reading comprehension and multi-sentence reasoning. 
These benchmarks provide a broader view of our models’ reasoning and generalization abilities. 
The results are summarized in Table~\ref{tab:challenging_benchmarks} and we can draw the following insights:
\begin{itemize}
    \item \textbf{1B scale:} The Zebra-Llama-1B (8MLA-8M2) achieves the best average score, outperforming the target Llama-3.2-1B-Instruct model with only 7.81\% of the KV cache.
    \item \textbf{3B scale:} The Zebra-Llama-3B models show only a 1.46\% and 0.55\% drop compared to the target model and X-EcoMLA, respectively, while achieving $21.3\times$ and $2\times$ KV cache compression. Even at 2.01\% KV cache, performance remains higher than both Mamba-In-Llama and Llamba.
    \item \textbf{8B scale:} The Zebra-Llama-8B variants exhibit a modest 7.1\% and 4.2\% performance gap relative to Llama-3.1-8B-Instruct and Mamba-In-Llama, respectively, despite $18.3\times$ and $9.1\times$ KV cache compression. The primary contributor to this gap appears to be the more challenging GSM8K task. We note that Mamba-In-Llama used a 70B teacher and 20B tokens, whereas our setup used an 8B teacher and 11B tokens, suggesting room for further improvement with additional training data.
    \item Across all model scales, \textbf{Zebra-Llama consistently outperforms Llamba}, confirming the effectiveness and scalability of our hybrid MLA–Mamba architecture.
\end{itemize}

\begin{table}[h!]
\centering
\resizebox{\textwidth}{!}{
\begin{tabular}{lcccccccc}
\toprule
\textbf{Model \& Setting} & \textbf{KV Size} & \textbf{Teacher} & \textbf{Tokens} & \textbf{Avg.} & \textbf{GSM8K (8-shot)} & \textbf{GPQA (Main)} & \textbf{GPQA (Diamond)} & \textbf{RACE} \\
\midrule
\multicolumn{9}{c}{\textit{1B Models}} \\
\midrule
Llama-3.2-1B-Instruct & 100\% & -- & -- & 32.44 & 38.51 & 26.79 & 26.26 & 38.18 \\
Mamba-In-Llama-1B & 50\% & 8B & 7B & 30.04 & 24.94 & 26.12 & 30.80 & 38.28 \\
X-EcoMLA-1B & 9.37\% & 8B & 7B & 31.30 & 38.06 & 21.43 & 26.26 & 39.43 \\
Llamba-1B & 0\% & 1B+70B & 8B & 27.67 & 22.82 & 25.00 & 25.25 & 37.61 \\
\textbf{Zebra-Llama-1B, 8MLA-8M2 (Ours)} & \textbf{7.81\%} & 8B & 7B & \textbf{32.64} & \textbf{41.09} & 25.67 & 25.25 & 38.56 \\
Zebra-Llama-1B, 4MLA-12M2 (Ours) & 3.91\% & 8B & 7B & 29.36 & 29.57 & 23.21 & 26.77 & 37.89 \\
\midrule
\multicolumn{9}{c}{\textit{3B Models}} \\
\midrule
Llama-3.2-3B-Instruct & 100\% & -- & -- & 42.32 & 70.89 & 29.24 & 28.28 & 40.86 \\
Mamba-In-Llama-3B & 50\% & 70B & 20B & 38.62 & 53.06 & 27.68 & 30.30 & 43.44 \\
X-EcoMLA-3B & 9.37\% & 8B & 7B & \textbf{41.93} & 65.28 & 27.46 & 30.30 & 44.69 \\
Llamba-3B & 0\% & 3B+70B & 10B & 34.35 & 47.76 & 22.77 & 26.77 & 40.10 \\
Zebra-Llama-3B, 14MLA-14M2 (Ours) & \textbf{4.69\%} & 8B & 9B & 41.70 & 62.77 & 27.23 & 29.80 & \textbf{46.99} \\
Zebra-Llama-3B, 6MLA-22M2 (Ours) & 2.01\% & 8B & 9B & 39.12 & 55.19 & \textbf{28.35} & \textbf{30.81} & 42.11 \\
\midrule
\multicolumn{9}{c}{\textit{8B Models}} \\
\midrule
Llama-3.1-8B-Instruct & 100\% & -- & -- & 47.42 & 78.17 & 35.49 & 31.31 & 44.69 \\
Mamba-In-Llama-8B & 50\% & 70B & 20B & 45.98 & 75.05 & 29.46 & 33.30 & 46.12 \\
X-EcoMLA-8B & 9.37\% & 8B & 7B & 44.78 & 70.81 & 29.69 & 30.30 & 48.33 \\
Llamba-8B & 0\% & 8B+70B & 12B & 38.28 & 57.62 & 28.35 & 26.77 & 40.38 \\
Zebra-Llama-8B, 16MLA-16M2 (Ours) & \textbf{5.47\%} & 8B & 11B & 44.05 & 68.16 & 29.02 & 30.30 & \textbf{48.71} \\
Zebra-Llama-8B, 8MLA-24M2 (Ours) & 2.73\% & 8B & 11B & 40.90 & 63.53 & 27.46 & 28.28 & 44.31 \\
\bottomrule
\end{tabular}
}
\caption{\small{\textbf{Additional Evaluation on Challenging Benchmarks.} 
Results are reported for GSM8K (8-shot), GPQA (Main and Diamond), and RACE. 
KV Size indicates the fraction of KV cache relative to the dense Llama baseline.}}
\label{tab:challenging_benchmarks}
\end{table}

}

\section{{MMLU Performance Discussion}}
 {When comparing the different size Zebra-Llama models to their corresponding base models, we observed that MMLU performance reaches approximately $84\%$ of the base model at the 1B scale, $86\%$ at the 3B scale, and 83\% at the 8B scale, which is pretty consistent performance across different sizes. We explain the potential reasons for the observed MMLU performance gap below:

\paragraph{MMLU Task Formatting Difficulty:} The observed gap in MMLU performance between hybrid models and pure Transformer baselines may stem from formatting sensitivity in MMLU’s multiple-choice structure. The study by~\cite{waleffe2024empirical} shows that state-space models (SSMs) like Mamba struggle with the standard MMLU format, which involves selecting a single letter (A/B/C/D) corresponding to the correct answer. While SSMs contain the same underlying knowledge as Transformers, they require more training to learn the task format, particularly the routing of information from multiple choices into a single output token—something Transformers handle more naturally via self-attention. This suggests that Zebra-Llama's hybrid use of Mamba/MLA layers may inherit some of this difficulty (inefficient decoding under MMLU’s standard prompting format), especially in tasks like MMLU that rely heavily on understanding structured input-output formatting, rather than open-ended generation or reasoning.

\paragraph{Importance of Training Data Selection}
\vspace{-4pt}

We believe that careful curation and selection of training datasets play a crucial role in enhancing model performance, particularly for knowledge-intensive benchmarks such as \textbf{MMLU}. 
Our \textit{Zebra-Llama} models were trained using the same datasets as \textit{Mamba-In-Llama} (our main initial baseline)—namely, \textbf{OpenHermes-2.5}~\cite{OpenHermes2_5}, \textbf{GenQA}~\cite{chen2024genqa}, and \textbf{Infinity-Instruct}~\cite{infinity_struct}. 
However, as highlighted in the \textit{Llamba} paper~\cite{bick2025llamba}, training on a carefully filtered version of \textbf{FineWeb-Edu}~\cite{penedo2024fineweb} leads to substantial improvements in MMLU scores following distillation.  

To further examine this effect, we re-trained a \textit{Llamba} model using our dataset configuration (without FineWeb-Edu).  As demonstrated in Table~\ref{tab:mmlu_dataset_comparison}, the MMLU score dropped from 38.11 (as reported in the original paper) to 27.35. 
In contrast, our \textit{Zebra-Llama} model—trained on the exact same dataset and using the same training pipeline—achieved a significantly higher MMLU score of \textbf{36.91}. 
This result suggests that \textit{Zebra-Llama} would likely surpass \textit{Llamba} even further if trained with similarly curated data. 
We view this as an exciting opportunity for future work, and encourage the broader community to explore the interplay between hybrid architectures and dataset quality in reasoning-intensive domains.
}

\begin{table}[h]
\centering
\resizebox{0.7\textwidth}{!}{
\begin{tabular}{lcc}
\toprule
\textbf{Model} & \textbf{Training Dataset} & \textbf{MMLU Score} \\
\midrule
Llamba (Original paper) & FineWeb-Edu (filtered) + OpenHermes-2.5 & 38.11 \\
Llamba (Our re-training) & OpenHermes-2.5 + GenQA + Infinity-Instruct & 27.35 \\
\textbf{Zebra-Llama (Ours)} & OpenHermes-2.5 + GenQA + Infinity-Instruct & \textbf{36.91} \\
\bottomrule
\end{tabular}
}
\caption{\small{\textbf{MMLU Comparison Across Training Datasets.}  
FineWeb-Edu filtering notably improves reasoning performance, and our Zebra-Llama model demonstrates strong robustness even without access to curated data.}}
\label{tab:mmlu_dataset_comparison}
\end{table}

\section{Evaluating \textit{Llamba} and \textit{Zebra-Llama} Under Consistent Training Conditions}
\label{appendix:llamba_comparison}
\vspace{-2pt}
 {

To provide a fair and controlled comparison between our Zebra-Llama and Llamba models, we re-implemented both \textit{Zebra-Llama} and \textit{Llamba} under identical training pipelines and datasets. 

We conducted this study at the \textbf{1B model scale} and Both models were trained using the three-stage pipeline introduced in the \textit{Llamba} paper. 
Since the filtered \textit{FineWeb-Edu} dataset is not publicly available, we used the same dataset configuration as in \textit{Zebra-Llama}, totaling approximately 6B tokens. 
We maintained the same token distribution across stages as in \textit{Llamba}: 225M tokens for Stage 1, 2.025B for Stage 2, and 3.75B for Stage 3. 
All other hyperparameters were kept identical to ensure consistency.

\begin{itemize}[leftmargin=10pt]
    \item \textbf{Target model:} Llama-3.2-1B-Instruct  
    \item \textbf{Teacher model:} Llama-3.2-1B-Instruct  
    \item \textbf{Initial learning rate:} 8e-5  
    \item \textbf{Batch size:} 96  
\end{itemize}

\noindent The results in Table~\ref{tab:llamba_vs_zebra} show that \textit{Zebra-Llama} surpasses \textit{Llamba} (pure SSM) across all eight tasks while using only \textbf{3.91\%} of the KV cache, highlighting the benefits of integrating efficient attention modules (MLA) with Mamba.

\begin{table}[h!]
\centering
\resizebox{\textwidth}{!}{
\begin{tabular}{lcccccccccc}
\toprule
\textbf{Model \& Setting} & \textbf{KV Size} & \textbf{Avg.} & \textbf{ARC} & \textbf{ARE} & \textbf{HS} & \textbf{MMLU} & \textbf{OBQA} & \textbf{PIQA} & \textbf{RA} & \textbf{WG} \\
\midrule
Llama-3.2-1B-Instruct & 100\% & 51.87 & 37.97 & 63.30 & 60.65 & 46.05 & 34.80 & 74.32 & 38.18 & 59.67 \\
Llamba-1B & 0\% & 47.23 & 34.98 & 62.12 & 54.41 & 27.35 & 34.60 & 71.82 & 34.83 & 57.70 \\
\textbf{Zebra-Llama-1B, 4MLA-12M2 (Ours)} & \textbf{3.91\%} & \textbf{49.32} & \textbf{37.12} & \textbf{62.67} & \textbf{55.54} & \textbf{37.10} & \textbf{35.80} & \textbf{72.20} & \textbf{35.50} & \textbf{58.64} \\
\bottomrule
\end{tabular}
}
\caption{\textbf{Controlled Comparison of \textit{Llamba} and \textit{Zebra-Llama} (1B scale).}  
Both models are trained with identical datasets and pipelines. \textit{Zebra-Llama} achieves higher average performance with over 25$\times$ KV cache compression.}
\label{tab:llamba_vs_zebra}

\end{table}

}

\newpage
\section*{NeurIPS Paper Checklist}

\begin{enumerate}

\item {\bf Claims}
    \item[] Question: Do the main claims made in the abstract and introduction accurately reflect the paper's contributions and scope?
    \item[] Answer: \answerYes{} 
    \item[] Justification: We summarized our contributions at the end of the introduction section in three main items: architecture, training and our results. 
    

\item {\bf Limitations}
    \item[] Question: Does the paper discuss the limitations of the work performed by the authors?
    \item[] Answer: \answerYes{} 
    \item[] Justification: The limitations of our work is discussed in the conclusion section. 

\item {\bf Theory assumptions and proofs}
    \item[] Question: For each theoretical result, does the paper provide the full set of assumptions and a complete (and correct) proof?
    \item[] Answer: \answerNA{} 
    \item[] Justification: Our paper is not a theory paper and we do not have theoretical assumption and proofs. 

    \item {\bf Experimental result reproducibility}
    \item[] Question: Does the paper fully disclose all the information needed to reproduce the main experimental results of the paper to the extent that it affects the main claims and/or conclusions of the paper (regardless of whether the code and data are provided or not)?
    \item[] Answer: \answerYes{} 
    \item[] Justification: We have all details of our experiments in the paper and appendix. We also will release the codes for reproducing our results when the paper is accpeted. 
    
    We provide detailed descriptions of our methodology, model architecture, training pipeline, and experimental setup in both the main paper and the appendix. This includes model configurations, training schedules, and evaluation procedures. While our code is not released at submission time, we commit to making them publicly available upon paper acceptance to further support reproducibility.

\item {\bf Open access to data and code}
    \item[] Question: Does the paper provide open access to the data and code, with sufficient instructions to faithfully reproduce the main experimental results, as described in supplemental material?
    \item[] Answer: \answerYes{} 
    \item[] Justification: We have described all the training and evaluation data in the paper and they are publicly available. Our codes for training are ready to be released when the paper is accepted.  

\item {\bf Experimental setting/details}
    \item[] Question: Does the paper specify all the training and test details (e.g., data splits, hyperparameters, how they were chosen, type of optimizer, etc.) necessary to understand the results?
    \item[] Answer: \answerYes{} 
    \item[] Justification: See the appendix for hyperparamters of our experiments. We have all the details of our experiments in the paper. 

\item {\bf Experiment statistical significance}
    \item[] Question: Does the paper report error bars suitably and correctly defined or other appropriate information about the statistical significance of the experiments?
    \item[] Answer: \answerNo{} 
    \item[] Justification: Due to the large scale of our models (ranging from 1B to 8B parameters), running multiple independent training runs to compute standard deviations or confidence intervals is computationally prohibitive. As a result, we do not report error bars. This is a common constraint in large model research, where the focus is typically placed on extensive ablations, benchmark diversity, and comparative baselines rather than repeated trials.


\item {\bf Experiments compute resources}
    \item[] Question: For each experiment, does the paper provide sufficient information on the computer resources (type of compute workers, memory, time of execution) needed to reproduce the experiments?
    \item[] Answer: \answerYes{} 
    \item[] Justification:     
    The paper reports detailed information about the computational resources used for training, including GPU type, training time, and hardware configuration. This information is presented both in the main results section and in the appendix. For instance, Table~\ref{appendix:model_structure} includes training time and model structure details. 
    
\item {\bf Code of ethics}
    \item[] Question: Does the research conducted in the paper conform, in every respect, with the NeurIPS Code of Ethics \url{https://neurips.cc/public/EthicsGuidelines}?
    \item[] Answer: \answerYes{} 
    \item[] Justification: We have carefully reviewed the NeurIPS Code of Ethics and confirm that our research complies with its guidelines. Our work does not involve human subjects, personally identifiable information, or any sensitive data. 

\item {\bf Broader impacts}
    \item[] Question: Does the paper discuss both potential positive societal impacts and negative societal impacts of the work performed?
    \item[] Answer: \answerYes{} 
    \item[] Justification: Our work focuses on improving the efficiency of large language models through post-training hybridization, making them more accessible for deployment in resource-constrained environments. The positive societal impacts include reducing the environmental and financial cost of training and serving LLMs, and enabling broader access to advanced language models, especially for smaller institutions or regions with limited computational infrastructure.
    
    
\item {\bf Safeguards}
    \item[] Question: Does the paper describe safeguards that have been put in place for responsible release of data or models that have a high risk for misuse (e.g., pretrained language models, image generators, or scraped datasets)?
    \item[] Answer: \answerNA{} 
    \item[] Justification: Our paper does not release any new datasets or pretrained models that pose a high risk of misuse. The research focuses on the methodology for constructing efficient hybrid language models using components from existing publicly available models. As such, it does not introduce new assets that require additional safeguards for responsible release.

\item {\bf Licenses for existing assets}
    \item[] Question: Are the creators or original owners of assets (e.g., code, data, models), used in the paper, properly credited and are the license and terms of use explicitly mentioned and properly respected?
    \item[] Answer: \answerYes{} 
    \item[] Justification:  
   All external assets used in our work—including pretrained models, datasets, and libraries—are properly cited in the paper. We used publicly available models and datasets under their respective open-source licenses. For each asset, we include appropriate references and URLs when applicable, and we ensure that all licenses and terms of use have been respected throughout our experiments.
   

\item {\bf New assets}
    \item[] Question: Are new assets introduced in the paper well documented and is the documentation provided alongside the assets?
    \item[] Answer: \answerYes{} 
    \item[] Justification: 
    We introduce new hybrid model variants and will release the codebase used for training and evaluation upon paper acceptance. The release will include documentation detailing the model architecture, training configurations, and usage instructions to support reproducibility and adoption. All assets will be shared under a permissive open-source license and accompanied by clear guidelines.

\item {\bf Crowdsourcing and research with human subjects}
    \item[] Question: For crowdsourcing experiments and research with human subjects, does the paper include the full text of instructions given to participants and screenshots, if applicable, as well as details about compensation (if any)? 
    \item[] Answer: \answerNA{} 
    \item[] Justification: 
    Our research does not involve any crowdsourcing or experiments with human subjects. All evaluations were conducted using automated benchmarks and publicly available datasets, without human annotation or feedback.

\item {\bf Institutional review board (IRB) approvals or equivalent for research with human subjects}
    \item[] Question: Does the paper describe potential risks incurred by study participants, whether such risks were disclosed to the subjects, and whether Institutional Review Board (IRB) approvals (or an equivalent approval/review based on the requirements of your country or institution) were obtained?
    \item[] Answer: \answerNA{} 
    \item[] Justification: Our paper does not involve crowdsourcing nor research with human subjects.

\item {\bf Declaration of LLM usage}
    \item[] Question: Does the paper describe the usage of LLMs if it is an important, original, or non-standard component of the core methods in this research? Note that if the LLM is used only for writing, editing, or formatting purposes and does not impact the core methodology, scientific rigorousness, or originality of the research, declaration is not required.
    \item[] Answer: \answerNA{} 
    \item[] Justification: While our work builds hybrid language models using components from existing pretrained LLMs, the usage of these models follows standard practices in the field (e.g., distillation, benchmarking, initialization). No LLMs were used in a novel or non-standard way as part of the methodology. Therefore, no declaration is required under the NeurIPS 2025 LLM policy.

\end{enumerate}


\end{document}